\documentclass[10pt,journal,compsoc]{IEEEtran}

\usepackage[utf8x]{inputenc}
\usepackage{amsmath}
\usepackage{amsfonts}
\usepackage{bm}
\usepackage{subcaption}
\usepackage{algorithm, algorithmicx, algpseudocode}

\usepackage{multirow}

\usepackage{graphicx}
\usepackage{ntheorem}
\newtheorem{theorem}{Theorem}
\newtheorem{lemma}[theorem]{Lemma}
\newtheorem{assumption}{Assumption}
\newtheorem{proposition}{Proposition}
\newtheorem{pf}{Proof}
\newtheorem{definition}{Definition}
\newtheorem*{pot}{Proof of Theorem 1}
\newtheorem*{pol}{Proof of Lemma 2}

\begin{document}

\title{Variational Counterfactual Prediction under Runtime Domain Corruption}

\author{Hechuan~Wen, Tong~Chen, Li~Kheng~Chai, Shazia~Sadiq, Junbin~Gao, Hongzhi~Yin\\
	\IEEEcompsocitemizethanks{
		\IEEEcompsocthanksitem H. Wen, T. Chen, S. Sadiq, and H. Yin, are with the School of Information Technology and Electrical Engineering, The University of Queensland, Brisbane, Australia\protect\\
		E-mail: h.wen@uq.edu.au, tong.chen@uq.edu.au, sadiq@itee.uq.edu.au, h.yin1@uq.edu.au 
		\IEEEcompsocthanksitem L. K. Chai is with Health and Wellbeing Queensland, Brisbane, Australia.\protect\\
		E-mail: LiKheng.Chai@hw.qld.gov.au
		\IEEEcompsocthanksitem J. Gao is with Business School, The University of Sydney, Sydney, Australia.\protect\\
		E-mail: junbin.gao@sydney.edu.au
}
\thanks{Hongzhi Yin is the corresponding author.}
}

\IEEEtitleabstractindextext{%
\begin{abstract}
To date, various neural methods have been proposed for causal effect estimation based on observational data, where a default assumption is the same distribution and availability of variables at both training and inference (i.e., runtime) stages. However, distribution shift (i.e., domain shift) could happen during runtime, and bigger challenges arise from the impaired accessibility of variables. This is commonly caused by increasing privacy and ethical concerns, which can make arbitrary variables unavailable in the entire runtime data and imputation impractical. We term the co-occurrence of domain shift and inaccessible variables \textit{runtime domain corruption}, which seriously impairs the generalizability of a trained counterfactual predictor. To counter runtime domain corruption, we subsume counterfactual prediction under the notion of domain adaptation. Specifically, we upper-bound the error w.r.t. the target domain (i.e., runtime covariates) by the sum of source domain error and inter-domain distribution distance. In addition, we build an adversarially unified variational causal effect model, named VEGAN, with a novel two-stage adversarial domain adaptation scheme to reduce the latent distribution disparity between treated and control groups first, and between training and runtime variables afterwards. We demonstrate that VEGAN outperforms other state-of-the-art baselines on individual-level treatment effect estimation in the presence of runtime domain corruption on benchmark datasets.
\end{abstract}

}

\maketitle

\IEEEdisplaynontitleabstractindextext

\IEEEpeerreviewmaketitle

\IEEEraisesectionheading{\section{Introduction}\label{sec:introduction}}
In predictive analytics, causal inference is increasingly important in guiding decision-making in high-stake domains, such as healthcare \cite{glass2013causal}, education \cite{cordero2018causal}, e-commerce \cite{wang2022causal}, etc. Normally, randomized control trial (RCT) is the gold standard for estimating the causal effect. Given that implementing RCTs is costly, time-consuming, and sometimes ethically intractable, various applications alternatively turn to use the passively collected observational data to perform causal inference in a data-driven fashion \cite{johansson2016learning,yao2021survey,gresele2022causal}. Denoting input variables as $\textbf{x}$, treatment as $t$, outcome as $y$, the observational dataset with $N$ samples $\{\boldsymbol(\textbf{x}_i, t_i, y_i)\}_{i=1}^{N}$ commonly does not satisfy the RCT standard due to \textit{unmeasured confounders} and \textit{selection bias}, which are two prominent challenges in causal inference. Specifically, the untestable unconfoundedness assumption assumes no unobserved confounders. Unfortunately, such an assumption cannot be satisfied in many cases, rendering the estimation erroneous \cite{rosenbaum2002overt,caliendo2008some}. Meanwhile, the selection bias between the treated and control groups causes the imbalanced covariate distributions, which could introduce undesirable spurious effect due to the imbalance \cite{yao2021survey}. In the extreme case, it can even violate the positivity assumption and result in non-identifiable causal effect \cite{jesson2020identifying}. Thus, this issue further weakens the correctness of causal effect estimation. 

An example to explain these two challenges is that, if only the rich can afford drug A while the poor have to use the cheaper drug B, then people's financial status could be a hidden confounder if unmeasured, resulting in invalid estimation. If measured, it causes selection bias, and the effectiveness of drug A and drug B cannot be validly compared based on the skewed distribution of variables due to people's financial status. By addressing either of the two challenges or both, several neural approaches \cite{shalit2017estimating,louizos2017causal,shi2019adapting} are made available for causal effect estimation with observational data.  
Despite a variety of methods that tackle distributional imbalance caused by the selection bias, such domain shifts are only restricted between the treated and control groups that are both used for training, where the runtime variables are assumed to be drawn from the same distribution as the training data. In fact, domain shift also widely exists between training and runtime data, e.g., when a model trained on one race is asked to perform predictions on a different minority race, and it challenges the generalizability of the trained model.


On top of that, the unavailable/missing variables and corresponding countermeasures are also largely understudied. For instance, real-world applications commonly have medical diagnostic models learned with high-quality open benchmarks, but in the deployment stage, not all end-users are able to provide the same set of variables due to accessibility issues (e.g., high-cost medical checks), privacy constraints (e.g., historical treatments), and ethical concerns (e.g., gender and race). In this paper, we refer to the co-existence of the shifted and unavailable variables in the inference data as \textit{runtime domain corruption}.

Runtime domain corruption can be interpreted as one step above observing domain/covariate shift during inference, where the model not only faces changed covariate distribution but also the ubiquitous missing values. 
In short, in our definition, runtime domain corruption is caused by the co-occurrence of domain shift and missing values. Compared with domain shift, runtime domain corruption more aggressively challenges the generalizability of the trained counterfactual prediction model, because variables deemed important in training might no longer be present during inference, and the domain-invariant patterns are unable to be mapped to those missing variables. Therefore, a high corruption rate of runtime variables can make the counterfactual predictor merely learned on full training data incur large generalization errors. Though one can consider discarding the unavailable variables in the training set, the reduced variables may lead to an underfitting issue. Also, for real-world deployment, it is impractical to assume prior knowledge on which variables are corrupted during runtime, especially considering the inaccessible variables can differ among individuals (e.g., users may choose to withhold different personal information). 


This work focuses on causal inference using the Neyman-Rubin potential outcome framework \cite{splawa1990application,rubin2005causal} under the runtime domain corruption circumstance. In this work, we aim to learn a robust, causal, and domain-invariant latent representation $\textbf{z}$ of variable $\textbf{x}$, for which the latent distributions across various domains are well-balanced to counter the aforementioned three challenges, i.e., unmeasured confounders, selection bias, and runtime domain corruption, simultaneously. Our main contributions are:
\begin{itemize}
\item We identify an important performance bottleneck for causal inference methods, namely runtime domain corruption that combines two largely unexplored yet important settings: domain shift and unavailable variables during runtime. In our paper, we propose the first systematic investigation of it for causal effect estimation.

\item We derive the upper bound of the generalization error by extending the in-sample causal inference to the corrupted out-of-sample scenario. To efficiently optimize the multiple Kullback-Leibler (KL) divergence terms in our VAE-based model, we propose a two-stage domain adaptation scheme, namely the Variational autoEncoder Generative Adversarial Network (VEGAN) for unifying multiple inter-domain distances. 

\item We compare VEGAN to state-of-the-art baselines for performing predictions on both in-sample covariates and out-of-sample, corrupted runtime covariates. The empirical results demonstrate our model's stronger robustness to runtime domain corruption.
\end{itemize}

\section{Related Work}

Back in time, researchers have been seeking to approach observational data-based causal inference from various perspectives. The re-weighting method, e.g., inverse probability weighting (IPW), uses the propensity score \cite{rosenbaum1983central, imbens2015causal} to mitigate the selection bias by re-weighting each unit's treatment outcome according to its estimated probability of being assigned a treatment. However, such a method strongly relies on the correctness of the estimated propensity score. To alleviate this strong dependency, the proposed doubly robust (DR) \cite{robins1994estimation} method considers the outcome regression together with IPW for re-weighing purposes. The DR method tries to secure the causal effect estimation with an additional ``insurance" that comes from the correctness of the outcome modelling which is in fact no one can assure. In addition to re-weighting, other methods such as the non-parametric tree-based model, e.g., BART \cite{chipman2010bart} combines the tree method and Bayesian inference. However, all the above-mentioned methods mainly focus on estimating the average treatment effect (ATE) and are not expressive enough to handle the high-dimensional dataset for individual-level estimations.

Nowadays, with the strong expressive power of deep learning \cite{bengio2013representation,lecun2015deep}, new algorithms are proliferating by leveraging the deep learning framework to learn the deconfounded latent representation on top of the observed covariates and model the personalized treatment effect. We relate our work to the representation learning branch in causal inference, which is overlapped with the domain adaptation field due to the unique counterfactual nature of estimating treatment effect. The TARNet \cite{shalit2017estimating} builds a shared feature extractor followed by a two-headed neural network to model the outcomes for each type of treatment separately. Its variants can incorporate the integral probability metric (IPM), e.g., Wasserstein distance \cite{vallender1974calculation}, and maximum mean discrepancy (MMD) \cite{gretton2012kernel}, to minimize the distance of the learned latent covariate distribution between treated and control groups to mitigate the selection bias. Following that, a variational autoencoder (VAE) framed CEVAE model \cite{louizos2017causal} emphasizes handling the confounding problem by building robust latent representation, and its performance is stated to be more robust than many previous methods. Dragonnet \cite{shi2019adapting} leverages the neural net-enhanced propensity estimation and the innovative targeted regularization for causal effect estimation to achieve an asymptotic consistent ATE estimator. In addition, other works such as GANITE \cite{yoon2018ganite} and DeepMatch \cite{kallus2020deepmatch} adopt generative adversarial network (GAN) \cite{goodfellow2014generative} and build their own designated GAN learning systems. Recently, many latent variable disentanglement methods, e.g., DR-CFR \cite{hassanpour2020learning}, TVAE \cite{vowels2020targeted}, TEDVAE \cite{zhang2021treatment}, are proposed to discover the disentanglement of the latent instrumental, risk, and confounding factors from the observed covariates to better capture the selection bias. The unique point of difference in our work is the additional consideration of the runtime domain corruption situation where the trained causal model's performance could dramatically decline when deployed to other environments. 

In addition, it is noted that \cite{jesson2020identifying} integrate the Monte Carlo Dropout \cite{gal2016dropout} method to the state-of-the-art neural methods and allow the upgraded models, e.g., BTARNET, BCEVAE, to estimate epistemic uncertainty in high-dimensional conditional average treatment effect (CATE) estimation, thus to inform the decision maker to be vigilant when making recommendations if the high uncertainty present. Their method only considers the domain shift between the treated and control groups during training, and it emphasizes making no treatment recommendation if the epistemic uncertainty exceeds a certain threshold. 
Hence, our work differs from it as we focus on more accurate treatment effect estimation when runtime domain corruption occurs during inference stage. It should also be noted that some existing works \cite{qu2009propensity,mayer2020doubly,berrevoets2022impute} have been proposed for treatment effect estimation with missing values, where the core is to leverage imputation algorithms to handle the missing values. Since runtime domain corruption also includes domain shift, the imputed target domain data could still deviate heavily from the source domain, rendering those methods inaccurate in such conditions. Furthermore, the imputation algorithm is not capable of imputing accurately when the number of missing values is large, it even becomes useless when the attributes are completely missing at the distribution level during the inference stage.

Lastly, we also relate our work to algorithmic fairness topics, e.g., disparate learning processes (DLPs), in which the ethically concerned, privacy-related features are not available or impermissible to be used during runtime \cite{lipton2018does,simons2021machine}. A similar approach to DLPs is a doubly-robust counterfactual prediction model with additional handling of the confounding problem during training \cite{coston2020counterfactual}. However, it differs from the common causal effect estimation as it assumes that one of the potential outcomes is a known constant for a binary treatment, and is hence inapplicable to the problem studied in this paper.

\section{Methodology \label{section:3}}

\subsection{Preliminaries}
For simplicity, we consider binary treatment $t$ of 1 or 0 to denote the treated group and the control group, respectively. The individual treatment effect (ITE) for a variable vector $\textbf{x}$ is defined as:
\begin{equation}\label{eq:true_effect}
   \tau(\textbf{x}) = \mathbb{E}[Y_{1} - Y_{0}| \textbf{x}],
\end{equation}
where $Y_{1}$ and $Y_{0}$ are the unobserved potential outcomes with treatment $t=1$ and $t=0$ respectively. As a common practice in causal inference research, to validly identify the true treatment effect $\tau(\textbf{x})$ of instance $\textbf{x}$, we make the following standard assumptions.

\begin{assumption}[Stable Unit Treatment Value Assumption]
   The SUTVA assumption \cite{imbens2015causal} states that: a) The potential outcomes for any unit do not vary with the treatment assigned to other units. b) For each unit, there are no different forms or versions of each treatment level, which leads to different potential outcomes.
\end{assumption}

\begin{assumption}[Unconfoundedness]
Treatment assignment is independent of the potential outcomes given the pre-treatment covariate $\textbf{x}$, i.e., $t\perp \{Y_{0}, Y_{1}\} | \textbf{x}$.
\end{assumption}

\begin{assumption}[Positivity]
    For every instance $\textbf{x}\in\mathcal{X}$, we have its corresponded treatment assignment mechanism $p(t|\textbf{x})$, such that $0<p(t=1|\textbf{x})<1$.
\end{assumption}

The assumptions further lead us to the proposition below.

\begin{proposition}[Identifiability]
The causal effect is identifiable if and only if the SUTVA, the unconfoundedness, and the positivity assumptions hold.
\end{proposition}

\begin{pf}
Under SUTVA and unconfoundedness, the ITE for instance $\textbf{x}$ is:
\begin{align}
\mathbb{E}[Y_{1}-Y_{0}|\textbf{x}] =& \mathbb{E}[Y_{1}|\textbf{x}] - \mathbb{E}[Y_{0}|\textbf{x}]\nonumber\\
=&\mathbb{E}[Y_{1}|X=\textbf{x},t=1]-\mathbb{E}[Y_{0}|X=\textbf{x},t=0]\nonumber\\
=&\mathbb{E}[y_{1}|X=\textbf{x},t=1]-\mathbb{E}[y_{0}|X=\textbf{x},t=0],
\end{align}
where $y_{1}$ and $y_{0}$ are the observed responses after the interventions $t=1$ and $ t=0$ have been taken, respectively. The last terms are identifiable as we assume $0<p(t=1|\textbf{x})<1$. The first equality is by the operation of expectation, the second equality is based on the unconfoundedness, and the third equality is by the expected value of the observed outcomes $\{y_{1}, y_{0}\}$ equals the unobserved potential outcomes $\{Y_{1}, Y_{0}\}$. 
\end{pf}


\subsection{Problem Definition}
Let $\Psi:\mathcal{X}\times\{0,1\}\rightarrow\mathbb{R}$ be the hypothesis, our goal is to build the treatment effect regression model $\Psi_{t}(\textbf{x},t)=\mathbb{E}[y_{t}|X=\textbf{x},T=t]$ with \textit{observed outcome $y_{t}$} based on the training data, which can accurately recover the treatment effect for test instance $\textbf{x}_{*}$, thus the causal effect for the test instance $\textbf{x}_{*}$ can be estimated as $\Psi_{1}-\Psi_{0}$. However, the \textit{untestable unconfoundedness} and \textit{selection bias} challenges arise when the observational dataset does not follow the RCT standard, making the trained models $\Psi_{1}$ and $\Psi_{0}$ unable to accurately reflect the true treatment outcomes for $\textbf{x}$.

We perceive the observed covariates of treated and control groups from the conventional domain shift perspective, in which covariate $\textbf{x}$ is a noisy measurement, normally less informative and more confounded \cite{griliches1986errors,maddala199617}, than the domain-invariant latent representation $\textbf{z}$. Therefore, the unconfoundedness changes from $t\perp \{Y_{0}, Y_{1}\} | \textbf{x}$ to $t\perp \{Y_{0}, Y_{1}\} | \textbf{z}$. In addition to the treated and control groups from the in-sample set, this paper uniquely considers causal effect estimation where the out-of-sample set is affected by runtime domain corruption. In what follows, we formally define the runtime domain corruption problem.

\begin{definition}[Runtime Domain Corruption]
We define each variable vector $\textbf{x}=[x_1, x_2,..., x_d]\in \mathbb{R}^{d}$ as a \textbf{non-zero} concatenation of categorical features (i.e., encodings) and numerical features. During training, all $d$ entries $x_s$, for $1\leq s\leq d$, are available and assigned corresponding values. Then, during inference, \textit{runtime domain corruption} occurs when: (1) the covariate distribution shifts in the test domain: $p_{test}(\textbf{x})\neq p_{train}(\textbf{x})$; and (2) each vector $\textbf{x}$ contains an arbitrary number of unavailable variables $x_{s'}$, for $1\leq s'\leq d$, which are all zeroed out by setting $x_{s'}=0$.
\end{definition}

\textbf{Rationale of zero-padding.} Specifically, during runtime, the unavailable features are not straightforwardly discarded when performing prediction, instead, we pad zeros to entries that correspond to missing variables such that the dimensionality is kept unchanged. It is worth noting that, during training, a non-zero property is maintained for every instance $\textbf{x}$ whose features are all available. This can be easily achieved via standard preprocessing steps, e.g., rescaling/normalization for numerical features, and using 1 and -1 to respectively represent relevant and irrelevant categorical features in multi-hot encodings. Thus, using zero-padding to mark unknown/corrupted variables during runtime is viable, because the semantics of zeros are exclusively reserved for the unknown status of variables. Also, zero-padding is a more feasible practice in real applications, as each runtime instance $\textbf{x}$ may have an arbitrary number and combination of attributes missing, rendering it impractical to train a specific latent feature extractor for each case. 
In contrast, zero-padding is a more flexible and scalable approach for learning domain-invariant latent representations with a shared feature extractor, where all zero-valued entries of $\textbf{x}$ will be filtered out during projection.

\subsection{Target Domain Error Upper Bound}

Shalit et al. \cite{shalit2017estimating} show that the accuracy metric for causal inference -- expected Precision in Estimation of Heterogeneous Effect (PEHE), denoted as $\epsilon_{\text{PEHE}}$, is upper-bounded by both the trained model error $\epsilon_{\text{F}}$ on factual outcomes and the distance between treated and control distributions, measured by integral probability metric (IPM). However, as their derived upper bound for $\epsilon_{\text{PEHE}}$ does not consider runtime domain corruption on out-of-sample variables, we fill this gap by deriving the bound in \textbf{Theorem} \ref{theorem:1}.

\begin{theorem}
\label{theorem:1}
\textit{Let $\phi:\mathcal{X}\rightarrow\mathcal{Z}$ be the invertible latent representation mapping function (a.k.a. feature extractor) with inverse $\Phi$. Let $\Psi:\mathcal{Z}\times\{0,1\}\rightarrow\mathbb{R}$ be the updated hypothesis that maps latent variables $\textbf{z}\in\mathcal{Z}$ to each treatment's outcome. Let $\mathcal{F} = \{f | f: \mathcal{Z}\rightarrow \mathbb{R}\}$ be a family of functions. The source domain is the observational data for treated and control groups, and the target domain is the runtime test/inference set with corrupted variables. We derive the upper bound of target domain error (i.e., generalization error) as\footnote{We follow some notation conventions set in \cite{shalit2017estimating} and \cite{johansson2020generalization}.}:}
\begin{equation}
\begin{split}
\epsilon^{tr}_{\text{PEHE}}
\leq2&\bigg{[}\epsilon_{\text{F}}^{t=1}+\epsilon_{\text{F}}^{t=0}+B_{\phi}\bigg{(}\text{IPM}_{\mathcal{F}}(\mathbb{P}^{t=1}_{\phi}, \mathbb{P}^{t=0}_{\phi})\\
+ &
\frac{1}{2}\text{IPM}_{\mathcal{F}}(\mathbb{P}^{tr}_{\phi}, \mathbb{P}^{sr}_{\phi})\bigg{)}\bigg{]},
\end{split}
\end{equation} \textit{where $\epsilon_{\text{F}}^{t}$ denotes the factual training error, $\mathbb{P}^{t}_{\phi}$ is the probability measure within treatment group $t$ in the training set, $\epsilon^{tr}_{\text{PEHE}}$ and $\epsilon^{sr}_{\text{PEHE}}$ respectively indicate the target and source domain errors, $\mathbb{P}^{tr}_{\phi}$ and $\mathbb{P}^{sr}_{\phi}$ are probability measures which respectively denote the covariate distributions in target domain and source domain, and $B_{\phi}$ is a bounded constant.} 

\end{theorem}

To prove \textbf{Theorem} \ref{theorem:1}, we first provide some preliminary definitions.

\begin{definition}
\label{def:ipm}
    \textit{Let $\mathcal{F} = \{f | f: \mathcal{Z}\rightarrow \mathbb{R}\}$ be a family of functions. The distribution distance measure -- integral probability metric (IPM) between the target and source distributions $\mathbb{P}^{sr}$ and $\mathbb{P}^{tr}$ over $\mathcal{Z}$ is defined as}:
\begin{equation}
\text{IPM}_{\mathcal{F}}(\mathbb{P}^{tr},\mathbb{P}^{sr}) = \sup_{f\in \mathcal{F}} \left|\int_{\mathcal{Z}}f(\textbf{z})(p^{tr}(\textbf{z})-p^{sr}(\textbf{z}))d\textbf{z}\right|.
\end{equation}
\end{definition}

\begin{definition}
\textit{Let $\phi:\mathcal{X}\rightarrow\mathcal{Z}$ be the latent mapping, and let $\Psi:\mathcal{Z}\times\{0,1\}\rightarrow\mathbb{R}$ be the updated hypothesis over the latent space $\mathcal{Z}$, the estimated ITE for variable $\textbf{x}$ is:}
\begin{equation}
    \hat{\tau}(\textbf{x}) = \Psi_{1}(\phi(\textbf{x}),1) - \Psi_{0}(\phi(\textbf{x}),0).
    \label{eq:estimation_def}
\end{equation}
\end{definition}

\begin{definition}
\label{def:pehe}
    \textit{The expected Precision in Estimation of Heterogeneous Effect (PEHE) of the causal model $\{\phi, \Psi\}$ with squared loss metric $L(\cdot, \cdot)$ is defined as:}
\begin{equation}
    \epsilon_{\text{PEHE}}(\phi,\Psi)=\int_{\mathcal{X}}L_{\phi, \Psi}(\textbf{x})p(\textbf{x})d\textbf{x},
    \label{eq:pehe_def}
\end{equation} where we denote $L(\hat{\tau}(\textbf{x}), \tau(\textbf{x}))$ as $L_{\phi, \Psi}(\textbf{x})$ for notation simplicity. The $\tau(\textbf{x})$ is the true treatment effect defined in Eq. \ref{eq:true_effect} and $\hat{\tau}(\textbf{x})$ is the estimated one defined in Eq. \ref{eq:estimation_def}.
\end{definition}

\begin{lemma}
\label{lemma:2}
\textit{ Let $\phi:\mathcal{X}\rightarrow\mathcal{Z}$ be the invertible latent representation mapping function with inverse $\Phi$. Let $\Psi:\mathcal{Z}\times\{0,1\}\rightarrow\mathbb{R}$ be the updated hypothesis. Let $\mathcal{F} = \{f | f: \mathcal{Z}\rightarrow \mathbb{R}\}$ be a family of functions. Assume we have $B_{\phi}>0$ s.t. $\frac{1}{B_{\phi}}L_{\phi,\Psi}(\Phi(\textbf{z}))\in \mathcal{F}$. The tightness of target domain error w.r.t. the source domain one is bounded by the distribution distance denoted by IPM:} 
\begin{equation}
\begin{split}
    &|\epsilon^{tr}_{\text{PEHE}} -\epsilon^{sr}_{\text{PEHE}}|\\=&\bigg{|}\int_{\mathcal{Z}}L_{\phi,\Psi}(\Phi(\textbf{z}))p^{tr}_{\phi}(\textbf{z})d\textbf{z} -  
    \int_{\mathcal{Z}}L_{\phi,\Psi}(\Phi(\textbf{z}))p^{sr}_{\phi}(\textbf{z})d\textbf{z}\bigg{|}\\
    \leq &B_{\phi}\text{IPM}_{\mathcal{F}}(\mathbb{P}^{tr}_{\phi},\mathbb{P}^{sr}_{\phi}),
\end{split}
\end{equation} \textit{where $\epsilon^{tr}_{\text{PEHE}}$ and $\epsilon^{sr}_{\text{PEHE}}$ indicate target domain error and source domain error respectively, $\mathbb{P}^{tr}_{\phi}$ and $\mathbb{P}^{sr}_{\phi}$ denote covariate distribution in target domain and source domain respectively, and $B_{\phi}$ is a bounded constant.}
\end{lemma}

\begin{pol}
    We denote the expected PEHE in Eq. \ref{eq:pehe_def} in target domain and source domain as $\epsilon_{\text{PEHE}}^{tr}$ and $\epsilon_{\text{PEHE}}^{sr}$ respectively, also $tr$ and $sr$ indicates the test set and training set where $p^{tr}(\textbf{x})\neq p^{sr}(\textbf{x})$ if domain corruption exists. 
    \begin{equation}
    \begin{split}
       |\epsilon_{\text{PEHE}}^{tr} & - \epsilon_{\text{PEHE}}^{sr}| 
        = \left|\int_{\mathcal{X}}L_{\phi,\Psi}(\textbf{x})p^{tr}_{\phi}(\textbf{x})d\textbf{x} - \int_{\mathcal{X}}L_{\phi,\Psi}(\textbf{x})p^{sr}_{\phi}(\textbf{x})d\textbf{x}\right|\\
        =&\left|\int_{\mathcal{Z}}L_{\phi,\Psi}(\Phi(\textbf{z}))p^{tr}_{\phi}(\textbf{z})d\textbf{z} - \int_{\mathcal{Z}}L_{\phi,\Psi}(\Phi(\textbf{z}))p^{sr}_{\phi}(\textbf{z})d\textbf{z}\right|\\
        =&\left|B_{\phi}\int_{\mathcal{Z}}\frac{1}{B_{\phi}}L_{\phi,\Psi}(\Phi(\textbf{z}))(p^{tr}_{\phi}(\textbf{z})-p^{sr}_{\phi}(\textbf{z}))d\textbf{z}\right|\\
        \leq&\left|B_{\phi}\sup_{f\in \mathcal{F}} \left|\int_{\mathcal{Z}}f(\textbf{z})(p^{tr}_{\phi}(\textbf{z})-p^{sr}_{\phi}(\textbf{z}))d\textbf{z}\right|\right|\\
        =&\left|B_{\phi}\text{IPM}_{\mathcal{F}}(\mathbb{P}^{tr}_{\phi}, \mathbb{P}^{sr}_{\phi})\right|\\
        =& B_{\phi}\text{IPM}_{\mathcal{F}}(\mathbb{P}^{tr}_{\phi}, \mathbb{P}^{sr}_{\phi}).
    \end{split}
\end{equation}
The first equality is by \textbf{Definition} \ref{def:pehe}, the second equality is by change of variable, the first inequality is by the premise that $\frac{1}{B_{\phi}}L_{\phi,\Psi}$ belongs to the function family $\mathcal{F}$, the fourth equality is by \textbf{Definition} \ref{def:ipm}, the last equality is by the property that IPM is non-negative.
\end{pol}

\begin{pot}
Under the conditions of \textbf{Lemma} \ref{lemma:2} and the auxiliary theorem 1 in \cite{shalit2017estimating}, thus conclude the proof of \textbf{Theorem} \ref{theorem:1}:
\begin{equation}
\begin{split}
\epsilon^{tr}_{\text{PEHE}}\leq&\epsilon^{sr}_{\text{PEHE}} + B_{\phi}\text{IPM}_{\mathcal{F}}(\mathbb{P}^{tr}_{\phi}, \mathbb{P}^{sr}_{\phi})\\
\leq2&[\epsilon_{\text{F}}^{t=1}+\epsilon_{\text{F}}^{t=0} \\
 +&B_{\phi}(\text{IPM}_{\mathcal{F}}(\mathbb{P}^{t=1}_{\phi}, \mathbb{P}^{t=0}_{\phi})+
\frac{1}{2}\text{IPM}_{\mathcal{F}}(\mathbb{P}^{tr}_{\phi}, \mathbb{P}^{sr}_{\phi}))],
\end{split}
\end{equation} where the first inequality is by \textbf{Lemma} \ref{lemma:2} and the second inequality is by the auxiliary theorem 1 from \cite{shalit2017estimating}. We align the function family $\mathcal{F}$ to the one used in \cite{shalit2017estimating}, as different choices of function family $\mathcal{F}$ will require different assumptions about the joint distribution $p(\textbf{z},t,y_{1},y_{0})$, the representation mapping function $\phi$, and the hypothesis $\Psi$. Thus, we share the same bounded constant $B_{\phi}$.
\end{pot}

In summary, the upper bound given in \textbf{Theorem} \ref{theorem:1} suggests that, to bring down the target domain error $\epsilon^{tr}_{\text{PEHE}}$ during runtime, we are essentially minimizing: (1) the prediction errors on observed outcomes; (2) the imbalance between treated and control groups; (3) the discrepancy between the training and test sets altogether. It guides our algorithm design in general for runtime causal inference. Note that if no domain corruption exists, which means $\mathbb{P}^{tr}_{\phi}=\mathbb{P}^{sr}_{\phi}$ and thus $\text{IPM}(\mathbb{P}^{tr}_{\phi}, \mathbb{P}^{sr}_{\phi})=0$, the runtime error becomes identical to the source domain error $\epsilon^{tr}_{\text{PEHE}} = \epsilon^{sr}_{\text{PEHE}}$.

\subsection{Variational Inference \label{section:4}}

Our solution is built upon the variational autoencoder (VAE). To start with, in this section we introduce the minimization of the factual error $\epsilon_{\text{F}}$, the distribution disparity between treated and control groups first, and between training and runtime domains afterwards.

\subsubsection{Evidence Lower Bound}
For modelling the \textit{observed treatment outcome $y$}, we use the maximum likelihood estimation (MLE) to approximate the parameters. For simplicity, log is commonly used to decompose the joint marginal likelihood $p(\textbf{y})$ into: 
\begin{equation}
\begin{split}
    \log p(\textbf{y})=& \sum_{k=1}^{N}\log p(y_k)\\
    =&\sum_{i=1}^{N_1}\log p(y_{i})+ \sum_{j=1}^{N_0}\log p(y_{j}),
    \label{eq:log-likelihood}
\end{split}
\end{equation} where $\textbf{y}=[y_1,y_2,\dots,y_N]$ is a vector hosting all $N$ samples' observations, $N = N_1 + N_0$, with $N_1$ and $N_0$ respectively denoting the number of samples in treated and control groups. Thus, to maximize the joint marginal log-likelihood of observing $\textbf{y}$, we can maximize each individual log-likelihood $\log p(y)$.

As we assume that there exists a latent representation $\textbf{z}$ and treatment $t$ that causally determine the observed treatment response $y_{t}$, i.e., $y_{t}\sim p(y|\textbf{z},t)$ in a probabilistic way, while the observed proxy $\textbf{x}$ does not have any causal relations but statistical correlations with $y$. Due to the potentially high dimensionality of $\textbf{z}$, the marginal likelihood $p(\textbf{y})$ is intractable. Here, we apply the variational methodology \cite{kingma2013auto} to our scenario to tackle $p(\textbf{y})$ by establishing an encoder network $\phi_t$ to learn the posterior latent representation $\textbf{z}_{t}\sim p_{\phi_t}(\textbf{z}|\textbf{x})$, and a decoder network $\Psi_{t}$ to estimate treatment response $y_{t}\sim p_{\Psi_t}(y|\textbf{z},t)$. According to the decomposed joint likelihood in Eq. \ref{eq:log-likelihood}, we can separately derive the the evidence lower bound ($\text{ELBO}_t$) for each of the treatment group $t$ in a similar manner used by \cite{kingma2013auto} as follows:
\begin{equation}
\begin{split}
    \sum_{i=1}^{N_t} \log p(y_{i}) \geq& \text{ELBO}_t\\
    =& \mathbb{E}_{\mathbb{P}_{\phi_t}}[\log p_{\Psi_t}(y|\textbf{z},t)] - D_{\text{KL}}(\mathbb{P}_{\phi_t}||\mathbb{Q}_{\textbf{z}}),
\end{split}
\label{eq:ELBO_t}
\end{equation} where $\mathbb{P}_{\phi_t}$ and $\mathbb{Q}_{\textbf{z}}$ are posterior and prior distributions respectively over the latent space $\mathcal{Z}$. $D_{KL}(\cdot)$ returns the Kullback-Leibler (KL) divergence between two distributions. As such, the task of maximizing the intractable $\log p(\textbf{y}_{t})$ can be indirectly solved by pushing up its associated $\text{ELBO}_t$, thus minimizing the factual error $\epsilon_{\text{F}}^{t}$. According to the decomposition in Eq. \ref{eq:log-likelihood}, our objective is to maximize the sum of two ELBOs for treated and control groups:
\begin{equation}
    \text{ELBO} = \sum_{t\in\{0,1\}}\text{ELBO}_{t}.
    \label{eq:ELBO}
\end{equation} 

It is worth noting that, our derived bound ELBO can be easily extended from our binary treatment setting to scenarios that involve multiple treatments.

\subsubsection{Treated/Control Domain Adaptation}

According to the second term in Eq. \ref{eq:ELBO_t}, for $t\in \{0,1\}$, we have both KL divergence terms that regularize the posterior distribution $\mathbb{P}_{\phi_t}$ and the prior distribution $\mathbb{Q}_{\textbf{z}}$, i.e., $ D_{\text{KL}}(\mathbb{P}_{\phi_1}||\mathbb{Q}_{\textbf{z}})$ and $D_{\text{KL}}(\mathbb{P}_{\phi_0}||\mathbb{Q}_{\textbf{z}})$. By pushing up the ELBO in Eq. \ref{eq:ELBO}, one can notice that both posteriors $\mathbb{P}_{\phi_1}$ and $\mathbb{P}_{\phi_0}$ are regularized to approach the same prior distribution $\mathbb{Q}_{\textbf{z}}$, e.g., standard normal distribution $\mathcal{N}(\textbf{0},\textbf{1})$. Thus, the domain adaptation (DA) for both groups can be naturally achieved to balance their latent distributions and counter selection bias by adjusting the priors using the VAE framework. It is worth noting that KL divergence is an unbounded asymmetric distribution distance measure \cite{kullback1951information} which does not belong to IPM, so we replace it with a bounded symmetric distribution similarity measurement in Section \ref{section:5} as a better approximation.

\subsubsection{Training/Runtime Domain Adaptation}

In addition to the DA across treated and control groups within the training set, we would also like to do DA between the entire training and runtime sets to minimize the tightness bound $B_{\phi}\text{IPM}_{\mathcal{F}}(\mathbb{P}^{tr}_{\phi}, \mathbb{P}^{sr}_{\phi})$ given in \textbf{Theorem} \ref{theorem:1} and thus alleviate runtime domain corruption. As such, for a well-trained model, we aim to make the out-of-sample performance as good as the in-sample performance, i.e., the out-of-sample results would not deviate from the in-sample ones drastically while keeping good in-sample performance.

Intuitively, if the VAE prediction framework is applied to the full runtime test set $\{(\textbf{x}_j^{tr}, t_j^{tr}, y_j^{tr})\}_{j=1}^{N'}$ on the target domain, one can end up with an objective to be maximized similar to the $\text{ELBO}_{t}$ presented in Eq. \ref{eq:ELBO_t} as follows: 
\begin{equation}
\Gamma_{\phi^{tr}_{t},\Psi^{tr}_{t}}=\mathbb{E}_{\mathbb{P}_{\phi_{t}}^{tr}}[\log p^{tr}_{\Psi_{t}}(y|\textbf{z},t)] - D_{\text{KL}}(\mathbb{P}_{\phi_t}^{tr}||\mathbb{Q}_{\textbf{z}}^{tr}).
\label{eq:testsetELBO}
\end{equation} 

However, the label $y^{tr}$ and treatments $t^{tr}$ are apparently unknown in practice, and such an objective cannot be optimized. Since the only available information is the runtime covariates which can be used to extract the domain-invariant representation from DA, the second term in Eq. \ref{eq:testsetELBO} can be utilised for such purpose with a mild modification.

Precisely, we alternatively walk around to minimize the KL divergence between the runtime posterior $\mathbb{P}_{\phi}^{tr}$ and the entire training set posterior $\mathbb{P}_{\phi}^{sr}$, namely $D_{\text{KL}}(\mathbb{P}_{\phi}^{tr}||\mathbb{P}_{\phi}^{sr})$, where $\phi$ is a shared feature extractor. Thus, to achieve the second-stage DA, our proposed ultimate evidence lower bound ($\text{ELBO}_{ulti}$) for the intractable joint log-likelihood $p(y)$ is: \begin{equation}
\begin{split}
    \log p(y) &\geq \text{ELBO}
    \\&\geq \text{ELBO}_{ulti}
    \\&= \text{ELBO} - D_{\text{KL}}(\mathbb{P}_{\phi}^{tr}||\mathbb{P}_{\phi}^{sr}).
    \label{eq:objective_ELBO}
\end{split}
\end{equation}

\subsection{Adversarial Learning \label{section:5}}

\begin{figure*}[t!]
    \centering
    \includegraphics[scale=0.9]{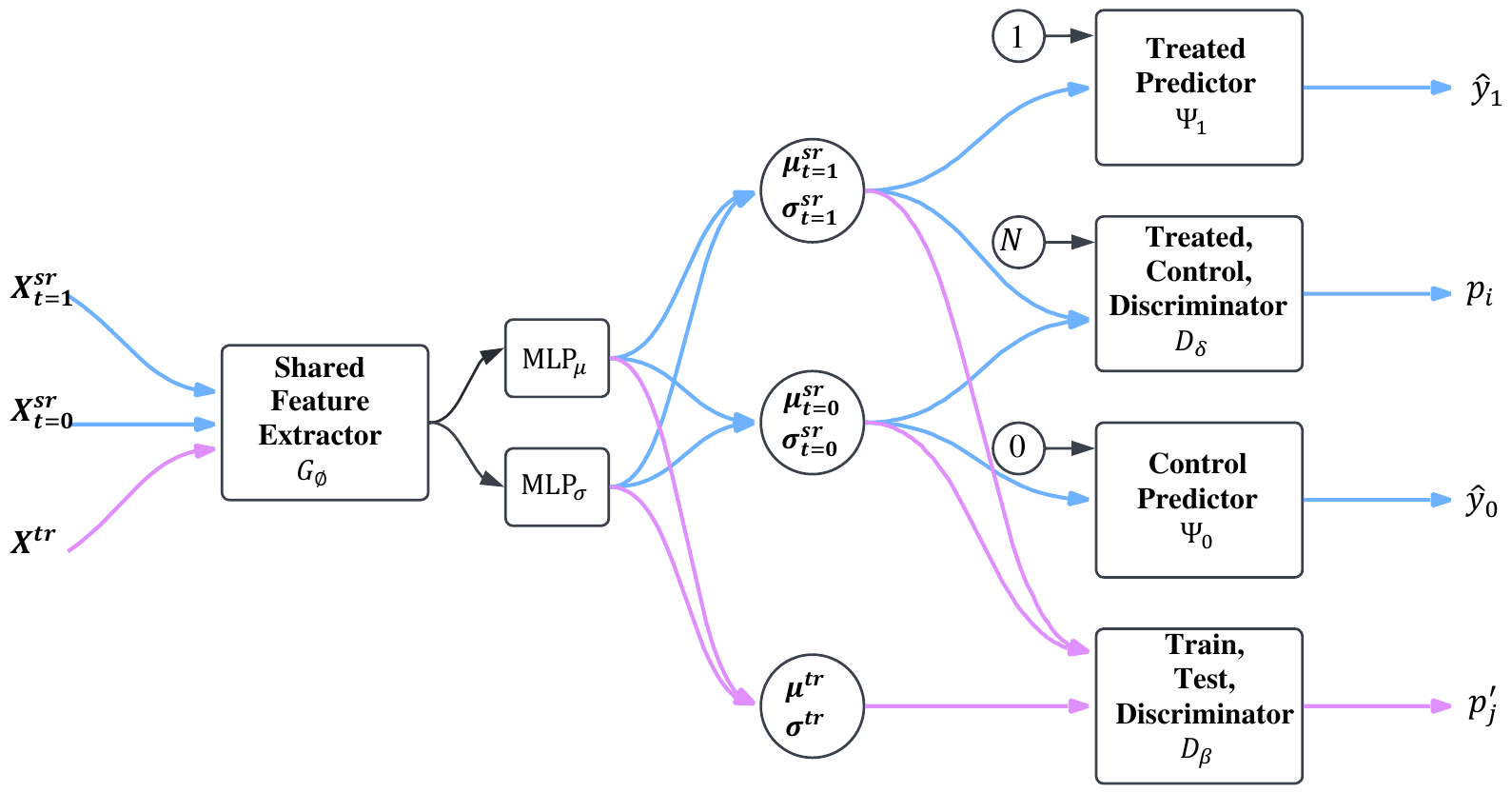}
    \caption{Unifying the KL-divergences by GAN. The black flows between $G_\phi$ and MLPs denote the internal link, as the feature extractor is two-headed modelling the mean and standard deviation of the latent distribution. The pink flows indicate second-stage domain adaptation between train and test sets.}
    \label{fig:two_step_da}
\end{figure*} 

Thus far, we have three $D_{\text{KL}}(\cdot)$ terms in our optimization objective: two from the in-sample treated/control groups, which align the posteriors to the same prior $\mathcal{N}(\textbf{0},\textbf{1})$, and the one from out-of-sample train/test adaptation that aligns the posteriors of training set and runtime set. As the direct calculation of KL divergence is computationally inefficient and may even be infeasible with high dimensional data \cite{nelken2006computing,moral2021computation}, we propose to implicitly minimize them and unify these terms into a compact generative adversarial network (GAN) \cite{goodfellow2014generative} shown in Figure \ref{fig:two_step_da}. Apart from that, optimizing the minimax game in GAN is equivalent to minimizing the Jensen-Shannon divergence \cite{goodfellow2014generative}, which is a bounded symmetric distribution similarity measurement \cite{e21050485}. Such technique resonates with adversarial variational Bayes introduced in \cite{mescheder2017adversarial}, while our motivation and implementation differ from theirs. Here, we present our $\textbf{V}$ariational auto$\textbf{E}$ncoder $\textbf{G}$enerative $\textbf{A}$dversarial $\textbf{N}$etwork runtime counterfactual regression model, coined as VEGAN. In what follows, we unfold the design details of VEGAN.

Firstly, we instantiate $\phi$, the shared feature extractor among $\textbf{x}^{sr}_{t=1}$, $\textbf{x}^{sr}_{t=0}$ and $\textbf{x}^{tr}$. It includes $G_{\phi}$ and the following two multi-layer perceptrons (MLPs) that map all the data from the original $\mathbb{R}^d$ into latent space $\mathbb{R}^l$. Due to the variational nature of the model, the $j$-th latent dimension of individual $i$, denoted by $\textbf{x}^{sr}_{i}$, is modelled by a Gaussian distribution with its dedicated mean $\mu_{ij}$ and variance $\sigma_{ij}^{2}$ as follows: \begin{equation}
    p_{\phi}^{sr}(\textbf{z}_i|\textbf{x}_i) = \prod_{j=1}^{l}\mathcal{N}(\mu_{ij},\sigma_{ij}^{2}),
\label{eq:p(z|x)}
\end{equation} 
where mean $\mu_{ij}$ and standard deviation $\sigma_{ij}$ are respectively the $j$-th element of latent representations $\bm{\mu}_i$ and $\bm{\sigma}_i$. In VEGAN, $\bm{\mu}_i, \bm{\sigma}_i\in \mathbb{R}^l$ are denoted as: 
\begin{equation}
\left\{
\begin{aligned}
 \bm{\mu}_i = \text{MLP}_{\mu}(G_{\phi}(\textbf{x}_{i}^{sr}))\\
 \bm{\sigma}_i = \text{MLP}_{\sigma}(G_{\phi}(\textbf{x}_{i}^{sr}))
\end{aligned} \right.,
\end{equation} which allows us to obtain the latent representation $\textbf{z}_i$ for the subsequent DAs and inference.

Secondly, for the treated/control group DA, we propose an adversarial way to implicitly reduce inter-domain distribution distance. In this regard, $G_{\phi}$ is essentially our generator for notation simplicity, and a discriminator $D_{\delta}$ is thus designed to pair up the generator to facilitate adversarial learning. The minimax game is designed as: the discriminator $D_{\delta}$ tries to differentiate the standard Gaussian sample $\textbf{n} \sim \mathcal{N}(\textbf{0},\textbf{1})$ from $\textbf{z}^{sr}$ learned from the training sample; in the meantime, feature extractor $G_{\phi}$ tries to update the latent representation $\textbf{z}^{sr}$ to make it indistinguishable from $\textbf{n}$. When an equilibrium state is reached, the treated and control domains are well adapted because both latent representations $\textbf{z}^{sr}$ fall in the same distribution. Thus, the two terms $D_{\text{KL}}(\mathbb{P}_{\phi_1}||\mathbb{Q}_{\textbf{z}})$ and $D_{\text{KL}}(\mathbb{P}_{\phi_0}||\mathbb{Q}_{\textbf{z}})$, are minimized in an adversarial way. As Figure \ref{fig:two_step_da} shows, the output $p_i = D_{\delta}(\textbf{w}_i)$ is the scalar probability of being a Gaussian sample, where $\textbf{w}_{i}=\eta_i\textbf{n}_{i}+(1-\eta_i)\textbf{z}^{sr}_i$ with $\eta_i \in \{1,0\}$ labelling the $i$-th sample from two buckets (1 for Gaussian samples, and 0 for training samples). Note that $\textbf{n}$ is resampled for every training instance $i\in \mathcal{I}$, where $\mathcal{I}$ is the collection of instances from the training set. In our supervised learning setting, we have the cross-entropy loss for the discriminator $D_{\delta}(\cdot)$: 
\begin{equation}
    l(\textbf{w}_{i})=\eta_i\log D_{\delta}(\textbf{w}_{i})+(1-\eta_i)\log(1-D_{\delta}(\textbf{w}_i)).
\end{equation}

Then, in an adversarial setting, the minimization of two KL-divergence terms for treated/control domain adaptation is replaced by the following: 
\begin{equation}
    \begin{split}
    &\min_{\phi}\max_{\delta}\mathbb{E}_{\mathcal{I}}[\eta_i\log D_{\delta}(\textbf{w}_{i})+(1-\eta_i)\log(1-D_{\delta}(\textbf{w}_i))] \\
    &\iff\min_{\phi}\max_{\delta}\mathbb{E}_{\mathcal{N}(\textbf{0}, \textbf{1})}[\log D_{\delta}(\textbf{n})]+\mathbb{E}_{\mathcal{I}}[\log(1-D_{\delta}(\textbf{z}_i^{sr}))].
       \label{eq:min-max-1}
    \end{split}
\end{equation}

\begin{algorithm}[t!]
\caption{Optimization of VEGAN}
\label{algorithm:VEGAN}
\begin{algorithmic}[1]
    \State $\textbf{Input:}$ Train set $\{\boldsymbol
(\textbf{x}_i^{sr}, t_i^{sr}, y_i^{sr})\}_{i=1}^{N}$, runtime variable set $\{({\textbf{x}}_{j}^{tr})\}_{j=1}^{N'}$ (optional), hyperparameters (e.g., learning rate $\alpha$), initialized neural networks' parameters $\{\phi, \Psi_1, \Psi_0, \delta, \beta\}$;
    \While{not converged}
    
    \State Sample a mini-batch $b_{m}\in\{b_{m}\}_{m=1}^{M}$ of size 
    \Statex \hspace{0.4cm} $|b_{m}|=|b_{m}^{t=1}|+|b_{m}^{t=0}|$, with $|b_{m}^{t=1}|=|b_{m}^{t=0}|$, from 
    \Statex \hspace{0.4cm} $\{\boldsymbol
(\textbf{x}_i^{sr}, t_i^{sr}, y_i^{sr})\}_{i=1}^{N}$. Sample equal size instances from 

	\Statex \hspace{0.4cm} standard Gaussian $\mathcal{N}(\textbf{0},\textbf{1})$;

    \State $\delta \leftarrow \delta - \alpha\frac{1}{|b_{m}|}\Big{(}\sum_{j=1}^{|b_{m}|}\nabla_{\delta}\log D_{\delta}(\textbf{n}_{j})$ 
    
    \Statex \hspace{1.4cm} $+\sum_{i=1}^{|b_{m}|}\nabla_{\delta}\log(1-D_{\delta}(\omega^{sr}_{i}))\Big{)}$;

    \If{runtime domain corruption exists}
    \State Randomly draw batch $b_{m}^{'}$ from $\{({\textbf{x}}_{j}^{tr})\}_{j=1}^{N'}$ with
    \Statex \hspace{0.9cm} size $|b_{m}^{'}|=|b_{m}|$;
    
    \State $\beta \leftarrow \beta - \alpha\frac{1}{|b_{m}|}\Big{(}\sum_{i=1}^{|b_{m}|}\nabla_{\beta}\log D_{\beta}(\omega^{sr}_{i})$
    
    \Statex  \hspace{1.0cm} $+\sum_{i=1}^{|b_{m}|}\nabla_{\beta}\log(1-D_{\beta}(\omega^{tr}_{i}))\Big{)}$;
    \EndIf

    \State $\phi\leftarrow \phi - \alpha \Big{(}\frac{1}{|b_{m}|}\sum_{i=1}^{|b_{m}|}\nabla_{\phi}\log D_{\delta}(\omega^{sr}_{i})$
    
    \Statex \hspace{1.0cm} $ + \frac{1}{|b_{m}|}\sum_{i=1}^{|b_{m}|}\nabla_{\phi}\log(1-D_{\beta}(\omega^{sr}_{i}))$
    
    \Statex \hspace{1.0cm} $+ \frac{1}{|b_{m}|}\sum_{i=1}^{|b_{m}|}\nabla_{\phi}\log D_{\beta}(\omega^{tr}_{i}) $
    
    \Statex \hspace{1.0cm} $+ \sum_{t\in\{0,1\}}\frac{1}{|b_{m}^{t}|}\sum_{i=1}^{|b_{m}^{t}|}\nabla_{\phi}\log p_{\Psi_t}(y_i|\omega_i,t_i)\Big{)}$,
    
    \State $\Psi_1\leftarrow \Psi_1 - \frac{\alpha}{|b_{m}^{t=1}|}\sum_{i=1}^{|b_{m}^{t=1}|}\nabla_{\Psi_1}\log p_{\Psi_1}(y_i|\omega_i,t_{i}=1)$,
    
    \State $\Psi_0\leftarrow\Psi_0 - \frac{\alpha}{|b_{m}^{t=0}|}\sum_{i=1}^{|b_{m}^{t=0}|}\nabla_{\Psi_0}\log p_{\Psi_0}(y_i|\omega_i,t_{i}=0)$;

    \EndWhile
\end{algorithmic}
\end{algorithm}

Analogously, for the train/runtime domain adaptation, we design another discriminator $D_{\beta}(\cdot)$ to form the second GAN system between $G_{\phi}(\cdot)$ and $D_{\beta}(\cdot)$, where $D_{\beta}(\cdot)$ predicts the probability $p_j'$ of the sample $j$, where $j\in \mathcal{J}$ is the collection of the test set, from the source domain (i.e., training set). The only difference from the first GAN system is that, it takes the training sample as real while the runtime sample is treated as fake. Thus, $D_{\text{KL}}(\mathbb{P}_{\phi}^{tr}||\mathbb{P}_{\phi}^{sr})$ is replaced by the following: \begin{equation}
    \begin{split}
      \min_{\phi}\max_{\beta}\mathbb{E}_{\mathcal{I}}[\log D_{\beta}(\textbf{z}^{sr}_i)]+\mathbb{E}_{\mathcal{J}}[\log(1-D_{\beta}(\textbf{z}^{tr}_j))].
       \label{eq:min-max-2}
    \end{split}
\end{equation}

Finally, to build the probabilistic model $p(y|\textbf{z},t)$, we model each of the treatment classes through two separate MLPs, namely $\Psi_{1}$ and $\Psi_{0}$ respectively, thus the general representation of modelling the observed outcome $y$ for individual $i$ is given as 
$p_{\Psi_t}(y_i|\textbf{z}_i, t_i) = \mathcal{N}(\hat{\mu}_{i},\hat{\sigma}^{2}_{i})$,
 where $\hat{\mu}_{t,i}=\Psi_{t_i}(\textbf{z}_{i}^{sr},t_i)$, and we follow \cite{louizos2017causal} to set $\hat{\sigma}^{2}_{i}=1$ for simplicity. 
 
In a nutshell, to promote a computationally efficient algorithm, we propose to minimize the following loss function $\mathcal{L}$ along with optimizing the minimax game together such that the $\text{ELBO}_{ulti}$ in Eq. \ref{eq:objective_ELBO} will be maximized: 
\begin{equation}
\begin{split}
\min_{\phi,\Psi_1,\Psi_0} &\max_{\delta,\beta}~\bigg{\{}\mathbb{E}_{\mathcal{N}}[\log D_{\delta}(\textbf{n})]+\mathbb{E}_{\mathcal{Z}^{sr}}[\log(1-D_{\delta}(\textbf{z}))] \\
&+
\mathbb{E}_{\mathcal{Z}^{sr}}[\log D_{\beta}(\textbf{z})]+\mathbb{E}_{\mathcal{Z}^{tr}}[\log(1-D_{\beta}(\textbf{z}))]+\mathcal{L}\bigg{\}},
\label{eq:final_objective}
\end{split}
\end{equation} where \begin{equation}
    \mathcal{L} = -\left(\mathbb{E}_{\mathbb{P}_{\phi_1}^{sr}}[\log p_{\Psi_1}(y|\textbf{z},t=1)]+\mathbb{E}_{\mathbb{P}_{\phi_0}^{sr}}[\log p_{\Psi_0}(y|\textbf{z},t=0)]\right).
\end{equation}

We summarize our VEGAN model optimization scheme in Algorithm \ref{algorithm:VEGAN}. Please note that the notation changed accordingly as the reparameterization trick $\textbf{z}=\omega(\boldsymbol{\mu}_{\phi},\boldsymbol{\sigma}^2_{\phi},\boldsymbol{\epsilon})$ \cite{kingma2013auto} is applied as a necessity to get the gradient $\nabla_{\phi}$ for the feature extractor $G_{\phi}$. Also, the original minimax game in Eq. \ref{eq:final_objective} is adjusted to the double minimization tradition for gradient descent.

\section{Experiments}

In this section, we evaluate the proposed VEGAN framework in dealing with the runtime domain corruption by answering the following research questions (RQs):
\begin{itemize}
    \item \textbf{RQ1}: How does VEGAN perform compared with other state-of-the-art models?
    \item \textbf{RQ2}: How effective is the proposed dual-stage DA in VEGAN?
    \item \textbf{RQ3}: Is VEGAN computationally efficient compared to other VAE-based models?
    \item  \textbf{RQ4}: As a classic solution to missing variables in prediction tasks, is data imputation on par with VEGAN's performance when handling runtime domain corruption?
    \item \textbf{RQ5}: Is our proposed second-stage plug-in applicable to other existing methods?
\end{itemize}

\subsection{Experimental Setup}

\begin{table}[b]
\centering
\caption{Tuned hyperparameters of VEGAN.}\label{table:modules}
\vspace{-0.2cm}
\setlength\tabcolsep{2.2pt}
\begin{tabular}{ccccc}
\hline
Module & \#Layers & \#Neurons & Learning Rate & Weight Decay\\
\hline
 $G_{\phi}$ & 3 & 100 &&\\
    $\Psi_1$ & 2 & 200 &&\\
    $\Psi_0$ & 2 & 200 &$10^{-3}$&$10^{-2}$\\
    $D_{\delta}$ & 2 & 100 &&\\
    $D_{\beta}$ & 2 & 100 &&\\
\hline
\end{tabular}
\end{table}

\subsubsection{Datasets and Domain Corruption Simulation} We utilize two popular semi-synthetic datasets in the causal inference literature, which are introduced below. 
\begin{itemize}
    \item \textbf{Infant Health and Development Program (IHDP) \cite{hill2011bayesian}}. The IHDP dataset contains 25 covariates and 747 samples, 
    assessing the effectiveness of early childhood interventions for low-birth-weight infants. To evaluate the causal model, the treatment outcomes are simulated according to \cite{hill2011bayesian}. In our test setting, seven privacy-related features are selected as target variables, i.e., \{momage, sex, twin, b.marr, cig, drugs, work\_dur\}, which are corrupted at different corruption levels (CLs), where the CL denotes the severity of the domain corruption ranging from 0\% to 100\%. While the rest 18 features remain unchanged. This is to mimic a typical runtime domain corruption scenario where individuals provide none or falsified privacy-related information for the trained model. We test $\text{CL}\in\{5\%, 12.5\%, 20\%, 33.3\%, 100\%\}$ on IHDP.
    \item \textbf{Atlantic Causal Inference Conference (ACIC) 2019 \cite{acic_2019}}. The ACIC 2019 dataset is a high-dimensional dataset of 200 covariates and 1,000 samples, which are drawn from publicly available data and the treatment outcomes are also simulated. In our test setting, since there is no clear definition of sensitive features, we treat all covariates as target variables for runtime domain corruption. As such, we test CL in $\{5\%, 12.5\%, 20\%, 33.3\%\}$, given that CL=100\% will wipe out all the covariates in ACIC.
\end{itemize}

Each of the dataset is randomly split with 3:1 ratio for train and test. 
As per our definition, runtime domain corruption entails both a shift in the covariate distribution and missing values in the test set. To simulate the distribution shift, for each target feature $x_{s}\in \textbf{x}_{i}$, we perform the following with the probability specified by each CL: (1) we add noise drawn from Gaussian distribution $\mathcal{N}(\Bar{\mu},0.1)$ to $x_{s}$ if $x_{s}$ is continuous; (2) we flip its value if $x_{s}$ is binary. To simulate the missing values, we scan each target feature $x_s$ and drop it (via zero-padding) with probability CL. The two corruption steps are performed independently on the same test set. 

\begin{table*}[t]
\parbox{.52\linewidth}{
\centering
\caption{$\sqrt{\epsilon_{\text{PEHE}}}$ of out-of-sample prediction on IHDP dataset with different corruption levels on private features.}\label{table:ihdp_ood}
\renewcommand{\arraystretch}{1.1}
\setlength\tabcolsep{1.8pt}
\begin{tabular}{llllll}
\hline
Model & 5\% & 12.5\% & 20\% &33.3\% & 100\%\\
\hline
TARNet &$1.725\pm .17$&$2.216\pm .26$&$2.455\pm .26$&$3.524\pm .42$&$5.845\pm .76$\\
    $\text{CFR}_{\text{WASS}}$&$\textbf{1.610}\pm \textbf{.15}$&$\textbf{2.070}\pm \textbf{.24}$ &$2.469\pm .27$&$3.503\pm .42$&$5.979\pm .85$\\
    SITE &$1.794\pm .19$&$2.136\pm .25$&$2.488\pm .27$&$3.427\pm .43$&$5.531\pm .70$ \\
    $\text{Dragonnet}_{\text{Base}}$ &$2.111\pm .24$&$2.272\pm .26$&$2.507\pm .28$&$3.190\pm .39$&$4.521\pm .60$\\
    $\text{Dragonnet}_{\text{TR}}$ &$2.023\pm .23$&$2.219\pm .25$&$2.516\pm .28$&$3.221\pm .40$&$4.760\pm .63$\\
    CEVAE &$3.074\pm .37$&$2.910\pm .32$&$2.987\pm .35$&$3.672\pm .46$&$4.164\pm .53$\\
    BTARNET & $2.375\pm .28$& $2.457\pm.29$& $2.566\pm.29$& $3.080\pm.39$& $4.229\pm.50$ \\
    BCEVAE & $2.506\pm.30$& $2.598\pm.32$& $2.783\pm.34$& $3.136\pm.39$& $4.170\pm.52$ \\
    TEDVAE & $2.232\pm.32$& $2.327\pm.34$& $2.582\pm.39$& $3.347\pm.41$& $4.453\pm.62$ \\
    $\text{VEGAN}$ &$1.720\pm .16$&$2.099\pm .23$&$\textbf{2.326}\pm \textbf{.25}$ &$\textbf{2.954}\pm \textbf{.35}$&$\textbf{3.918}\pm \textbf{.47}$\\
\hline
\end{tabular}}
\hfill
\parbox{.45\linewidth}{
\centering
\caption{$\sqrt{\epsilon_{\text{PEHE}}}$ of out-of-sample prediction on ACIC dataset with different corruption levels on all features.}\label{table:acic_ood}
\renewcommand{\arraystretch}{1.1}
\setlength\tabcolsep{1.8pt}
\begin{tabular}{lllll}
\hline
Model & 5\% & 12.5\% & 20\% &33.3\% \\
\hline
TARNet &$0.813\pm .05$&$0.738\pm .04$&$0.798\pm .05$&$0.754\pm .04$\\
    $\text{CFR}_{\text{WASS}}$ &$0.730\pm .04$&$0.655\pm .03$&$0.738\pm .04$&$0.644\pm .04$\\
    SITE &$1.482\pm .08$&$1.361\pm .08$&$1.365\pm .09$&$1.586\pm .10$\\
    $\text{Dragonnet}_{\text{Base}}$ &$2.250\pm .02$&$2.192\pm .02$&$2.142\pm .02$&$2.063\pm .02$\\
    $\text{Dragonnet}_{\text{TR}}$ &$2.517\pm .02$&$2.456\pm .02$&$2.406\pm .02$&$2.342\pm .02$\\
    CEVAE &$0.646\pm .02$&$0.629\pm .02$&$0.594\pm .02$&$0.581\pm .03$\\
    BTARNET & $0.705\pm.02$& $0.652\pm.02$& $0.653\pm.02$& $0.671\pm.02$ \\
    BCEVAE & $0.495\pm.02$& $0.507\pm.02$& $0.492\pm.02$& $0.532\pm.01$ \\
    TEDVAE & $0.736\pm.03$& $0.702\pm.03$& $0.775\pm.03$& $0.674\pm.03$ \\
    $\text{VEGAN}$ &$\textbf{0.490}\pm \textbf{.01}$&$\textbf{0.493}\pm \textbf{.01}$&$\textbf{0.471}\pm \textbf{.01}$&$\textbf{0.455}\pm \textbf{.00}$\\
\hline
\end{tabular}
}
\end{table*}

\subsubsection{Baselines and Evaluation Metrics}
We compare VEGAN with nine causal inference baselines as the following:
\begin{itemize}
    \item \textbf{TARNet} \cite{shalit2017estimating} is a base deep learning framework with a shared feature extractor and two decoders modelling the treated and control effects, respectively.
    \item \textbf{$\text{CFR}_{\text{WASS}}$} \cite{shalit2017estimating} is a variant of TARNet, with a latent distribution balancing regularization (Wasserstein distance) to overcome the confounding bias introduced by the imbalance between the treated and control groups.
    \item \textbf{CEVAE} \cite{louizos2017causal} is a variational autoencoder framework which focuses on modelling the robust latent variable to handle the confounding bias from a probabilistic perspective.
    \item \textbf{SITE} \cite{yao2018representation} explores the importance of the local similarity preservation as a constraint to improve ITE estimation, and proposes a deep representation learning method to help preserve the local similarity and balance data distribution altogether.
    \item \textbf{$\text{Dragonnet}_{\text{Base}}$} \cite{shi2019adapting} is based on TARNet, but it additionally provides an end-to-end procedure for predicting propensity score to adjust the confounding bias when estimating the treatment effects.
    \item \textbf{$\text{Dragonnet}_{\text{TR}}$} \cite{shi2019adapting} is built on top of $\text{Dragonnet}_{\text{Base}}$, the updated model introduces the novel targeted regularization based on the non-parametric estimation theory, which provides an asymptotic property with a suitable downstream estimator.
    \item \textbf{BTARNET} \cite{jesson2020identifying} enhances the decoders of TARNet with Monte Carlo dropout technique to quantify the uncertainty when estimating the treatment effect.
    \item \textbf{BCEVAE} \cite{jesson2020identifying} takes CEVAE as a base model, and incorporates the Monte Carlo dropout into its generative network for uncertainty quantification.
    \item \textbf{TEDVAE} \cite{zhang2021treatment} is a latent variable disentangle model based on a three-headed variational autoencoder, which tries to learn the disentangled latent instrumental, risk, and confounding factors, respectively, from the observed covariates.
\end{itemize}

\begin{table}[b]
\parbox{.45\linewidth}{
\centering
\caption{$\epsilon_{\text{CATE}}$ and $\sqrt{\epsilon_{\text{PEHE}}}$ of in-sample counterfactual prediction on IHDP dataset.}\label{table:ihdp_id}
\renewcommand{\arraystretch}{1.1}
\setlength\tabcolsep{1.2pt}
    \begin{tabular}{ccc}
\hline
Model & $\epsilon_{\text{CATE}}$ & $\sqrt{\epsilon_{\text{PEHE}}}$ \\
\hline
TARNet &0.253&1.033\\
    $\text{CFR}_{\text{WASS}}$ &0.265&$\textbf{0.992}$\\
    SITE &0.256&1.016\\
    $\text{Dragonnet}_{\text{Base}}$ &0.342&1.591\\
    $\text{Dragonnet}_{\text{TR}}$ &0.357&1.524\\
    CEVAE &0.329&2.604\\
    BTARNET &0.335&1.913\\
    BCEVAE &0.312&2.060\\
    TEDVAE &0.341&2.210\\
    $\text{VEGAN}$ &\textbf{0.239}&1.394\\
\hline
\end{tabular}}
\hspace{0.5cm}
\parbox{.45\linewidth}{
\centering 
\caption{$\epsilon_{\text{CATE}}$ and $\sqrt{\epsilon_{\text{PEHE}}}$ of in-sample counterfactual prediction on ACIC dataset.}\label{table:acic_id}
\renewcommand{\arraystretch}{1.1}
\setlength\tabcolsep{1.2pt}
\begin{tabular}{ccc}
\hline
Model & $\epsilon_{\text{CATE}}$ & $\sqrt{\epsilon_{\text{PEHE}}}$ \\
\hline
    TARNet &0.214 &0.808 \\ 
    $\text{CFR}_{\text{WASS}}$ &0.216 &0.708 \\
    SITE &0.293 &1.383\\
    $\text{Dragonnet}_{\text{Base}}$ &0.186 & 2.244\\
    $\text{Dragonnet}_{\text{TR}}$ &0.183 & 2.503\\
    CEVAE &0.286 &0.702 \\
    BTARNET & 0.225&0.704\\
    BCEVAE & 0.165&0.502\\
    TEDVAE & 0.202&0.651\\
    $\text{VEGAN}$ &\textbf{0.147} & \textbf{0.476}\\
\hline
\end{tabular}}
\end{table}

\begin{figure*}[t]
    \centering
    \includegraphics[width=0.95\textwidth]{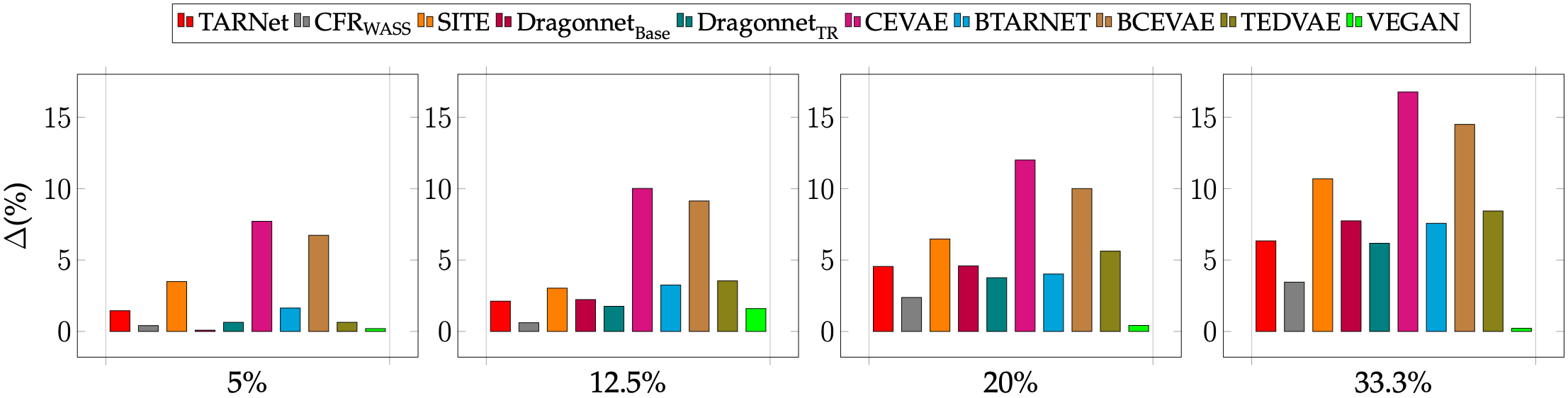}
\vspace{-0.2cm}
    \caption{Performance volatility $\Delta$ of all models on ACIC dataset under different CLs.}
    \label{fig:volatility}
\vspace{-0.7cm}
\end{figure*}

\subsubsection{Implementation}

Our model is implemented with PyTorch \cite{paszke2019pytorch}. The hyperparameters are tuned according to the models' performance on validation set. Our tuned hyperparameters are shown in Table \ref{table:modules}, respectively. All the experiments are conducted with RTX-3090 on Ubuntu 22.04 LTS platform where GPU training is enabled, otherwise the 12th Gen Intel i7-12700K 12-Core 20-Thread CPU is used.

\subsection{Performance Evaluation (RQ1)}

\subsubsection{Out-of-Sample Prediction under Runtime Domain Corruption}

\textbf{IHDP Dataset}. For predictions on the corrupted, out-of-sample instances, we conduct the tests on the test set with five corruption ratios $\text{CL}\in\{5\%, 12.5\%, 20\%, 33.3\%, 100\%\}$. Notably, $\text{CL}=100\%$ represents an extreme case where all the seven sensitive features are completely inaccessible during runtime and only the remaining 18 variables are available for prediction. As Table \ref{table:ihdp_ood} demonstrates, $\text{VEGAN}$ yields the second best performance when the domain corruption is relatively restrained, and obtains the highest accuracy after the corruption ratio increases to and beyond 20\%. The best baseline is $\text{CFR}_{\text{WASS}}$ when CL is low, but it overfits the training set significantly and thus does not generalize to a higher domain corruption level, while $\text{VEGAN}$ is more robust to the stronger corruption on IHDP's private variables. 

\textbf{ACIC}. Since there is no clear definition for all 200 features on ACIC 2019 dataset, we allow the corruption to take place for all the features in ACIC dataset with a ratio of $\text{CL}\in\{5\%, 12.5\%, 20\%, 33.3\%\}$. With this, we can mimic situations where individuals can withhold an arbitrary combination of variables in privacy-sensitive applications. Note, that we omit $\text{CL}=100\%$ in ACIC dataset as it will set all variables to zero and thus make any predictions infeasible. As a result, VEGAN outperforms all the other models for out-of-sample prediction as shown in Table \ref{table:acic_ood}.

\subsubsection{In-Sample Prediction without Runtime Domain Corruption}

Besides the out-of-sample prediction under the run-time corruption, we also investigate the traditional in-sample inference, where there no corruption happens, i.e., there is neither distribution shift nor missing variables. Table \ref{table:ihdp_id} and \ref{table:acic_id} show the in-sample prediction results on both CATE and ITE estimation, for which our model performs the best in estimating CATE while staying competitive for ITE estimation on the IHDP dataset, and outperforms all the other models on the ACIC dataset. 

\subsubsection{Volatility Analysis} 
It is noted that when the domain corruption level climbs, the fluctuations of prediction errors are small in magnitude on ACIC 2019 dataset. To better quantify the advantage of VEGAN under domain corruption on ACIC, we analyse the deviation ($\Delta$) of each model's performance between in-sample and corrupted prediction tasks in Figure \ref{fig:volatility}, i.e., $\Delta = 100\% \times |\epsilon_{\text{in-sample}}-\epsilon_{\text{corrupted}}|/\epsilon_{\text{in-sample}}$. $\Delta$ quantifies the instability of the model, as we commonly rely on models obtained with the training set and prefer lower generalization errors. All models become more volatile as CL increases, while $\text{VEGAN}$ maintains an excellent stability with only 0.22\% variation at corruption level $33.3\%$ and achieves the best accuracy in terms of $\sqrt{\epsilon_{\text{PEHE}}}$.

\subsection{Effectiveness of Second-Stage DA (RQ2)}

\begin{figure}[t]
\centering
\begin{subfigure}{\linewidth}
\hspace{1.2cm}\includegraphics[width=0.7\linewidth]{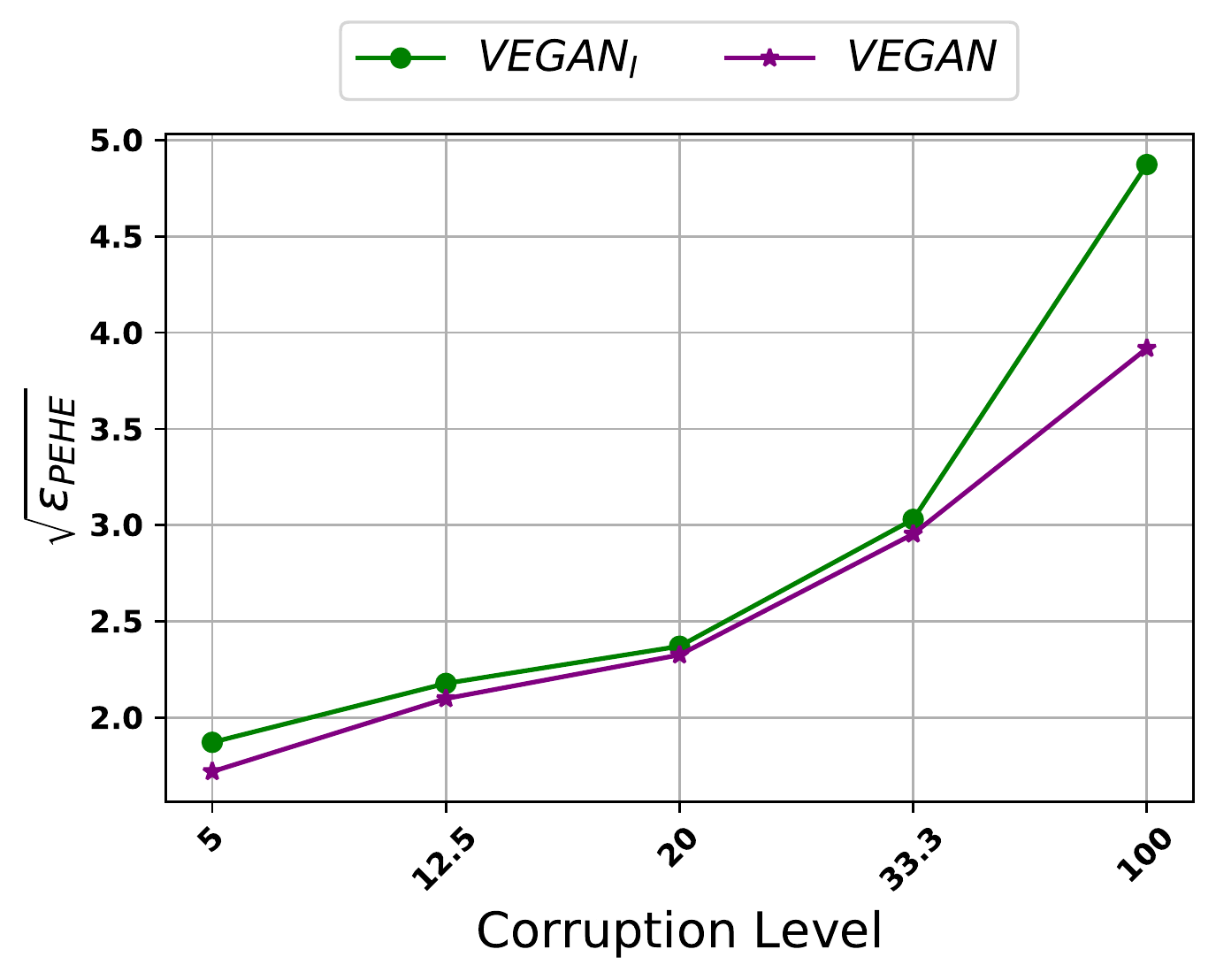}
\vspace{-0.2cm}
    \caption{IHDP}
\end{subfigure}%

\begin{subfigure}{\linewidth}
\hspace{1.2cm}\includegraphics[width=0.7\linewidth]{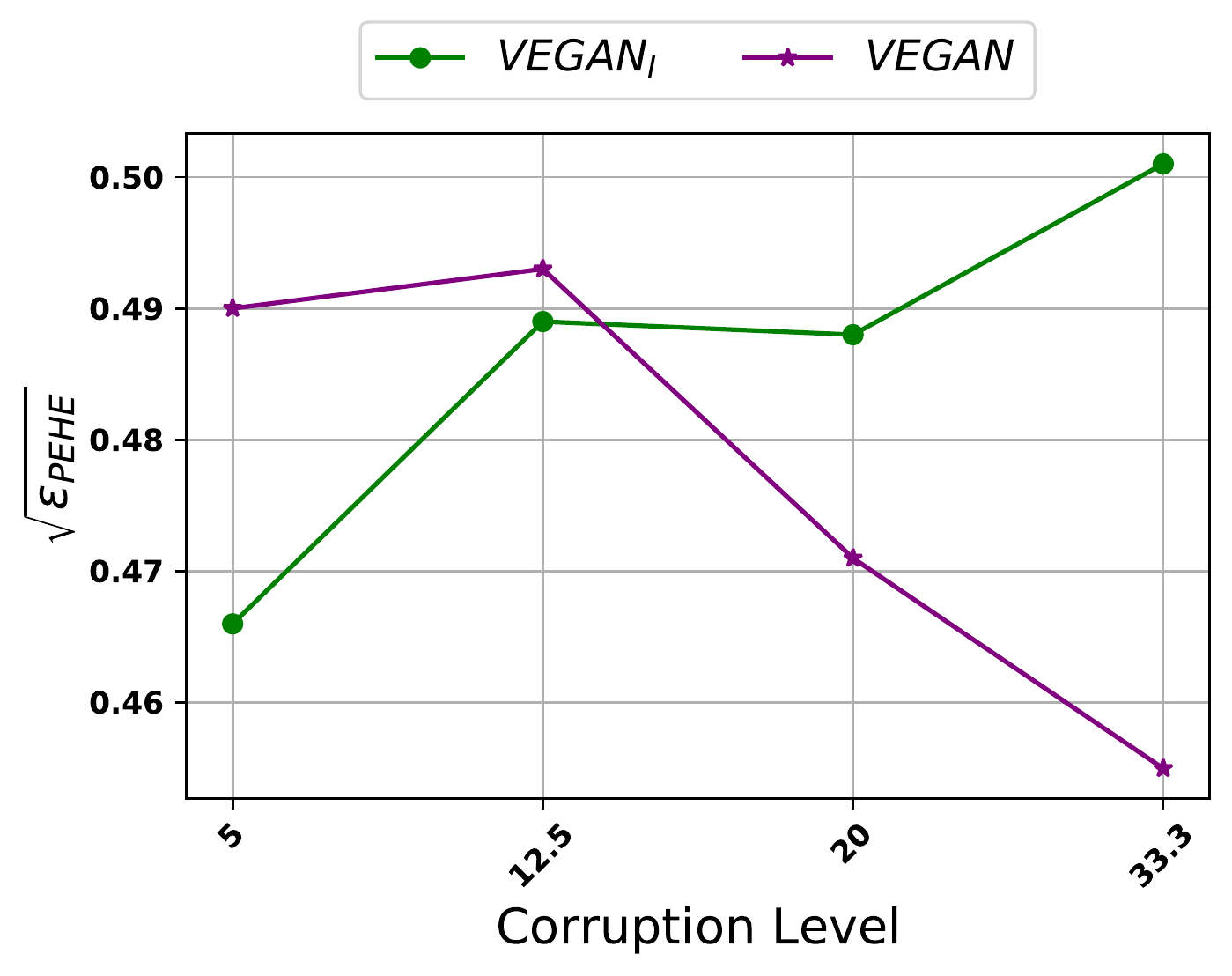}
\vspace{-0.2cm}
    \caption{ACIC}
    
\end{subfigure}%
\caption{Ablation study on second-stage adversarial plug-in on VEGAN framework.}
\label{fig:ablation}
\end{figure}

As $\text{VEGAN}$'s main highlight is the second-stage adversarial DA as a plug-in component, we conduct an ablation study to compare the performance of VEGAN and VEGAN\textsubscript{I} on both datasets, where VEGAN\textsubscript{I} is a degraded version with the second-stage DA removed. The results in Figure \ref{fig:ablation} indicate that, when CL is low, both two models are comparable. However, when CL goes higher, the advantage of the second-stage plug-in becomes significant. Thus, with our proposed second-stage DA, $\text{VEGAN}$ is shown to have higher generalization ability than $\text{VEGAN}_{\text{I}}$ across different scenarios.

\subsection{Computational Efficiency \& Stability of VEGAN (RQ3)}

\begin{figure}[t]
\centering
\includegraphics[width=0.3\textwidth]{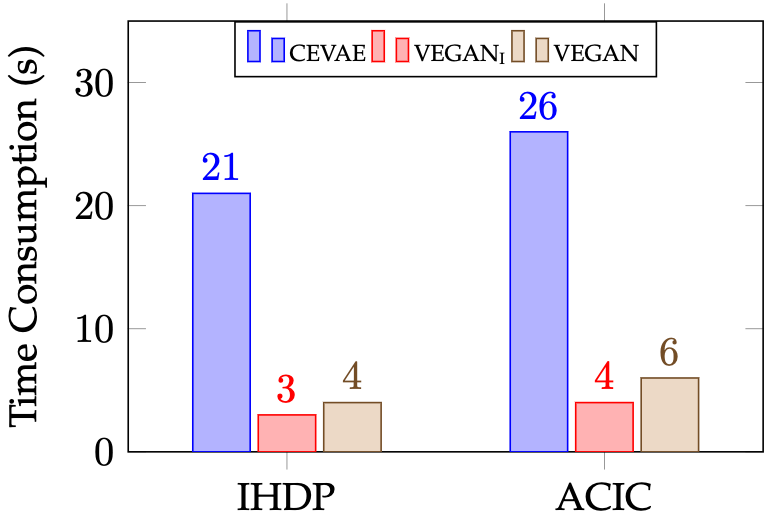}
\vspace{-0.2cm}
    \caption{Computational efficiency comparison between CEVAE and our VEGAN framework.}
    \vspace{-0.2cm}
    \label{fig:comp_time}
\end{figure}

\begin{figure}[t]
\centering
\includegraphics[scale=0.27]{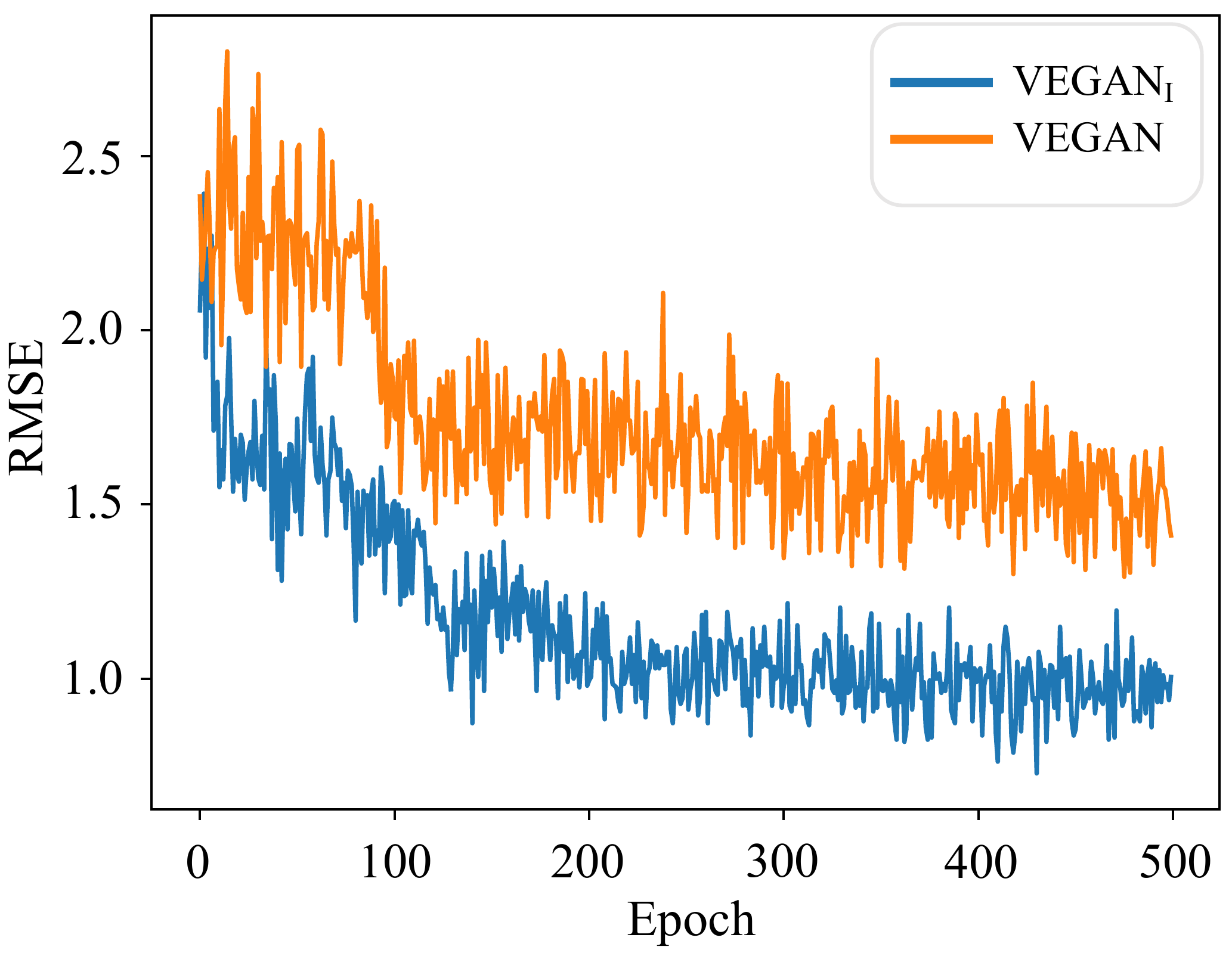}
\vspace{-0.2cm}
    \caption{Convergence of $\text{VEGAN}_{\text{I}}$ and $\text{VEGAN}$ on ACIC dataset in terms of RMSE.}
    \label{fig:convergence}
\end{figure}

One core motivation for utilizing GAN to replace the straightforward KL divergence optimization is to preserve training efficiency under high dimensionality. Hence, we further test VEGAN's efficiency by comparing its training time (in seconds) per 100 epochs with CEVAE and VEGAN\textsubscript{I}. To ensure a fair comparison, the tests are performed on 12th Gen Intel i7-12700K 12-Cores 20-Threads CPU on Ubuntu 22.04 LTS. Figure \ref{fig:comp_time} shows that $\text{VEGAN}_{\text{I}}$, which can be viewed as an amplified version of CEVAE with the introductions of GAN, has significantly faster training speed (over 6$\times$ speedup). Furthermore, the introduction of our second-stage adversarial DA in $\text{VEGAN}$ is still able to maintain high computational efficiency, witnessed by over 4$\times$ speedup over CEVAE.

As GAN is known to be unstable during training, We provide the stability analysis in terms of prediction loss convergence and equilibrium status between feature extractor and discriminators. Figure \ref{fig:convergence} shows the models' convergence on root mean square error (RMSE) during training. It is noted that a higher corruption level brings more challenges in the adversarial training, but we observe an equilibrium state in the majority of the cases. Taking the more challenging 33.3\% CL in ACIC dataset as an example, both discriminators (for treated/control and training/runtime adaptations) can quickly converge to the equilibrium by returning an average binary cross-entropy loss of 0.69, which means the discriminators are completely deceived by the feature extractors, and always give 0.5 probability for the samples from each of the groups. As such, training VEGAN in an adversarial setting is completely attainable. 


\subsection{Comparison with Imputation Method (RQ4)}

As imputation is a natural choice to handle missing values, we test the effectiveness of VEGAN against data imputation methods on ACIC's corrupted out-of-sample test sets. Specifically, we implement the imputation algorithm MICE \cite{van2000multivariate}, which has been widely adopted in treatment effect estimation \cite{mayer2020doubly,berrevoets2022impute}. We denoted imputation-enhanced models with ``*'' in Table \ref{table:acic_imp}. The results indicate that, when the corruption rate is low, using imputation is generally helpful for slightly increasing the prediction performance compared to Table \ref{table:acic_ood}, but the improvements remain marginal and less significant compared with VEGAN. In short, data imputation has very limited benefits under the domain corruption setting. Furthermore, in scenarios where an attribute is completely missing for all instances, it is infeasible to impute this attribute based on its distribution within existing test samples for prediction.


\begin{table}[t]
\caption{$\sqrt{\epsilon_{\text{PEHE}}}$ of out-of-sample prediction on ACIC dataset for baselines with imputation enhanced.}\label{table:acic_imp}
\renewcommand{\arraystretch}{1.1}
\setlength\tabcolsep{2.5pt}
\begin{tabular}{lllll}
\hline
Model & 5\% & 12.5\% & 20\% &33.3\% \\
\hline
TARNet* &$0.807\pm .04$&$0.728\pm .04$&$0.783\pm .04$&$0.763\pm .04$\\
    $\text{CFR}_{\text{WASS}}\text{*}$ &$0.722\pm .04$&$0.640\pm .03$&$0.710\pm .04$&$0.679\pm .04$\\
    SITE* &$1.322\pm .08$&$1.210\pm .07$&$1.358\pm .08$&$1.268\pm .08$\\
    $\text{Dragonnet}_{\text{Base}}$* &$2.232\pm .02$&$2.162\pm .02$&$2.097\pm .02$&$2.117\pm .02$\\
    $\text{Dragonnet}_{\text{TR}}$* &$2.485\pm .02$&$2.407\pm .02$&$2.343\pm .02$&$2.393\pm .02$\\
    CEVAE* &$0.626\pm .02$&$0.601\pm .02$&$0.577\pm .02$&$0.568\pm .03$\\
    BTARNET* & $0.714\pm.02$& $0.647\pm.02$& $0.658\pm.02$& $0.689\pm.02$ \\
    BCEVAE* & $0.522\pm.02$& $0.497\pm.01$& $0.491\pm.02$& $0.511\pm.02$ \\
    TEDVAE* & $0.752\pm.02$& $0.712\pm.02$& $0.765\pm.02$& $0.682\pm.02$ \\
    $\text{VEGAN}$ &$\textbf{0.490}\pm \textbf{.01}$&$\textbf{0.493}\pm \textbf{.01}$&$\textbf{0.471}\pm \textbf{.01}$&$\textbf{0.455}\pm \textbf{.00}$\\
\hline
\end{tabular}
\end{table}

\begin{table*}[t]
\centering
\caption{ITE error $\sqrt{\epsilon_{\text{PEHE}}}$ and variation $\Delta$(\%) of out-of-sample prediction on ACIC dataset with different corruption levels when the second stage domain adaptation is applied to TARNet.}\label{table:acic_plugin}
\renewcommand{\arraystretch}{1.1}
  \begin{tabular}{lllllllll}
    \hline
    \multirow{2}{*}{Model} &
      \multicolumn{2}{c}{5\%} &
      \multicolumn{2}{c}{12.5\%)} &
      \multicolumn{2}{c}{20\%)} &
      \multicolumn{2}{c}{33.3\%} \\
      \cline{2-9}
      & $\sqrt{\epsilon_{\text{PEHE}}}$ & $\Delta$(\%) & $\sqrt{\epsilon_{\text{PEHE}}}$ & $\Delta$(\%) & $\sqrt{\epsilon_{\text{PEHE}}}$ & $\Delta$(\%) & $\sqrt{\epsilon_{\text{PEHE}}}$ & $\Delta$(\%)\\
    \hline
    TARNet & $0.813\pm .05$ & \textbf{1.45} & $0.738\pm .04$ & 2.12 & $0.798\pm .05$ & 4.55 & $0.754\pm .04$ & 6.34 \\
    $\text{TARNet}_{+}$ & $\textbf{0.540}\pm \textbf{.01}$ & 2.21 & $\textbf{0.545}\pm \textbf{.02}$ & \textbf{0.29} & $\textbf{0.526}\pm \textbf{.02}$ & \textbf{1.90} & $\textbf{0.554}\pm \textbf{.02}$ & \textbf{2.29} \\
    \hline
  \end{tabular}
\end{table*}

\subsection{Applicability of Second-Stage DA to Other Baselines (RQ5)}
To demonstrate the applicability of our proposed second-stage adversarial DA plug-in to other state-of-the-arts, we study its compatibility with the most representative baseline TARNet. The experiments are conducted using the ACIC dataset, and the results are presented in Table \ref{table:acic_plugin}. We denote the TARNet with adversarial plug-in as $\text{TARNet}_{+}$. As the results suggest, there is a transferable benefit to the other baseline with our proposed second-stage adversarial plug-in when the corruption level becomes higher, the benefit of the second-stage domain adaptation will be enlarged. When the adversarial plug-in is in use, it effectively helps TARNet reduce prediction risks under runtime domain corruption as the volatility of the $\text{TARNet}_{+}$ is stabilized at around 2\%.

\section{Conclusion}
This paper formalizes the runtime causal inference problem under domain corruption, where novel strategies are proposed to counter the imbalance between treated and control groups and the inter-domain discrepancy between training and inference domain. We further adopt adversarial learning to replace the direct calculation of KL-divergence to improve computational efficiency. For our proposed approach VEGAN framework with second-stage domain adaptation, its performance exceeds other state-of-the-arts under the runtime domain corruption setting in semi-synthetic and full-synthetic benchmark datasets. In addition, the second-stage adversarial plug-in is demonstrated as applicable to the off-the-shelf models to reduce generalization errors.

\section*{Acknowledgement}
This work is supported by the Australian Research Council under the streams of Industrial Transformation Training Centre (No. IC200100022), Future Fellowship (No. FT210100624), Discovery Project (No. DP190101985), and Discovery Early Career Researcher Award (No. DE230101033).

\vskip -15pt plus -1fil
\begin{IEEEbiography}[{\includegraphics[width=1in,height=1.25in,clip,keepaspectratio]{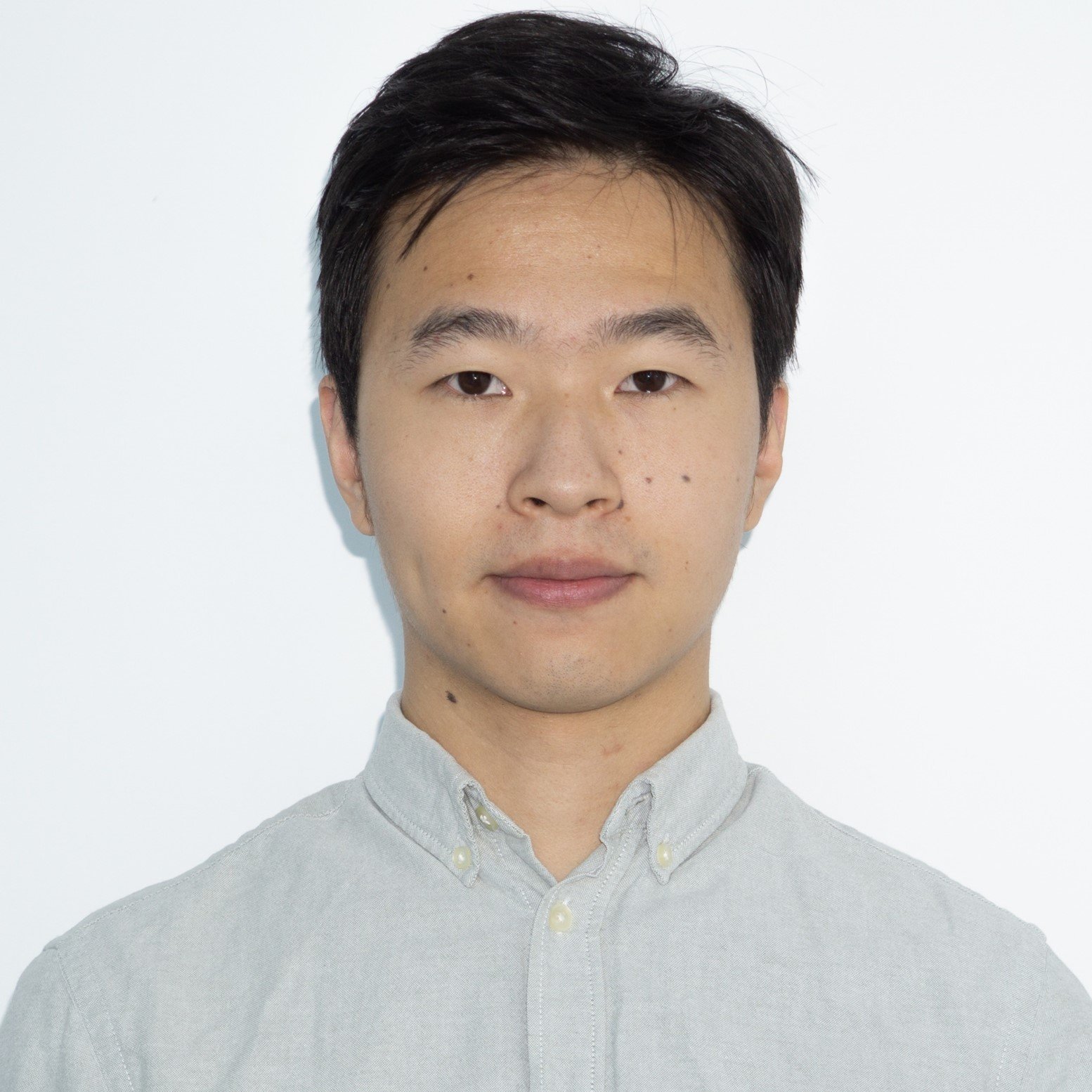}}]{Hechuan Wen}
	 completed his B.Eng. degree at Nanjing University of Aeronautics and Astronautics in 2019, and M.Com. degree at The University of Sydney in 2021. Currently, he is a Ph.D. candidate at the School of Information Technology and Electrical Engineering, the University of Queensland. His research interests include causal inference, domain adaptation, uncertainty quantification and variance reduction.
\end{IEEEbiography}
\vskip -15pt plus -1fil
\begin{IEEEbiography}[{\includegraphics[width=1in,height=1.25in,clip,keepaspectratio]{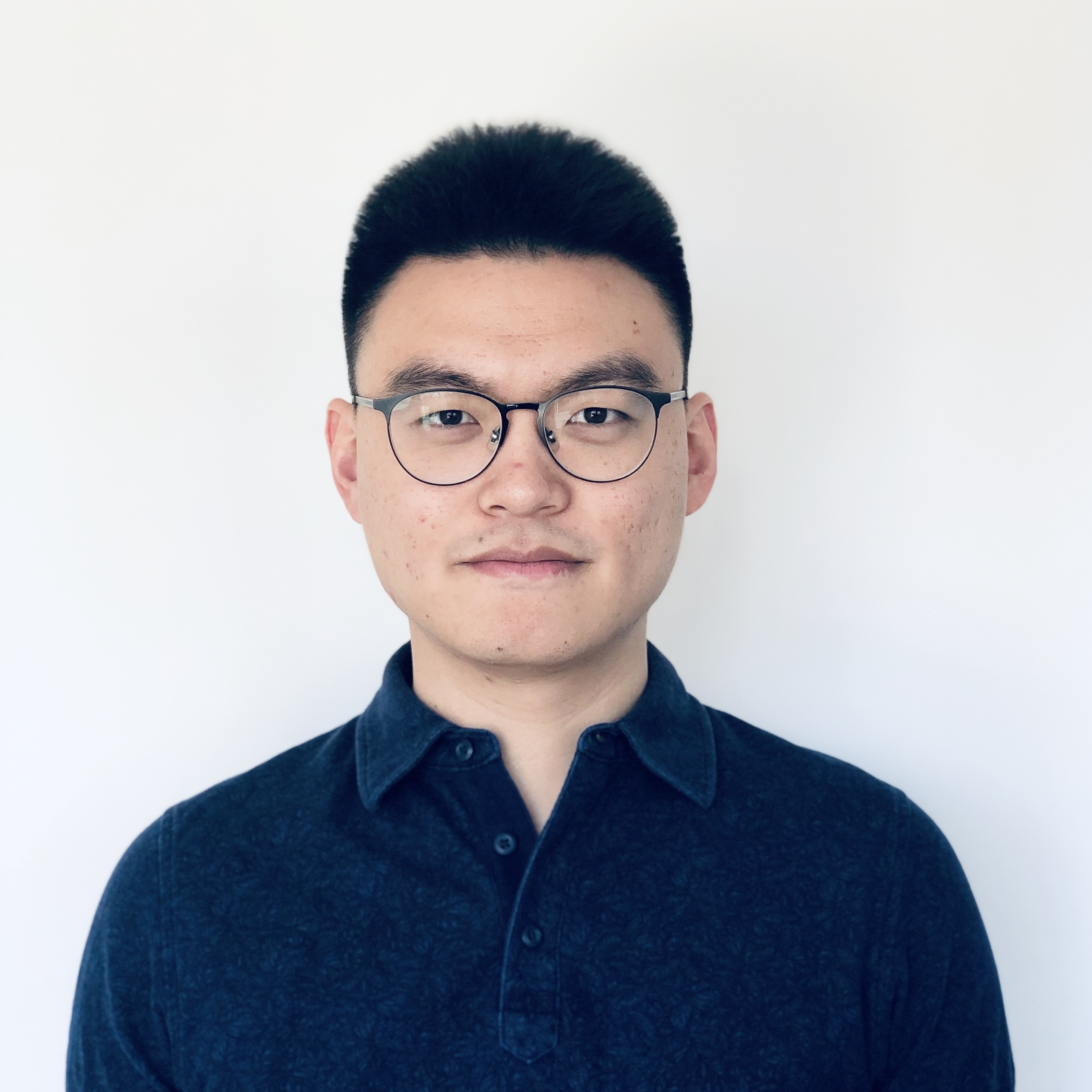}}]{Tong Chen}
received his PhD degree in computer science from The University of Queensland in 2020. He is currently a lecturer and ARC DECRA Fellow with the Data Science research group, School of Information Technology and Electrical Engineering, The University of Queensland. His research interests include data mining, recommender systems, user behavior modelling and predictive analytics. 
\end{IEEEbiography}
\vskip -15pt plus -1fil
\begin{IEEEbiography}[{\includegraphics[width=1in,height=1.25in,clip,keepaspectratio]{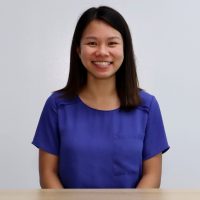}}]{Li Kheng Chai}
is a Dietitian, Research Fellow at Health and Wellbeing Queensland, Honorary Fellow at University of Queensland, and Social Media Editor (Honorary) of European Journal of Clinical Nutrition. Dr Chai completed her PhD in Nutrition and Dietetics at the University of Newcastle and has experience in undergraduate teaching and supervision, health promotion, implementation science and health service research related to children's health and obesity prevention and intervention.
\end{IEEEbiography}
\vskip -15pt plus -1fil
\begin{IEEEbiography}[{\includegraphics[width=1in,height=1.25in,clip,keepaspectratio]{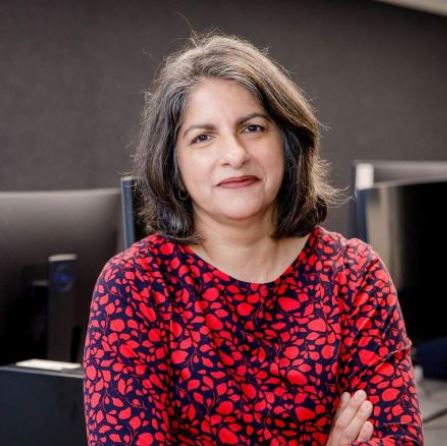}}]{Shazia Sadiq}
is currently a professor with the School of Information Technology and Electrical Engineering, The University of Queensland. She has authored or coauthored more than 100 peer-reviewed publications in high-ranking journals, such as Information Systems Journal, VLDBJ, TKDE, and also main conferences, such as SIGMOD, ICDE, ER, BPM, ICIS, and CAiSE. Her main research interests include innovative solutions for Business Information Systems that span several areas, including business process management, governance, risk and compliance, and information quality and use.
\end{IEEEbiography}
\vskip -15pt plus -1fil
\begin{IEEEbiography}[{\includegraphics[width=1in,height=1.25in,clip,keepaspectratio]{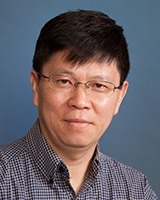}}]{Junbin Gao}
Junbin Gao received the B.Sc. degree in computational mathematics from the Huazhong University of Science and Technology (HUST), Wuhan, China, in 1982, and the Ph.D. degree from the Dalian University of Technology, Dalian, China, in 1991.
From 1982 to 2001, he was an Associate Lecturer, a Lecturer, an Associate Professor, and a Professor with the Department of Mathematics, HUST. He was a Senior Lecturer and a Lecturer in computer science with the University of New England, Armidale, NSW, Australia, from 2001 to 2005. He was a Professor in computer science with the School of Computing and Mathematics, Charles Sturt University, Australia. He is currently a Professor of big data analytics with the University of Sydney Business School, The University of Sydney, Sydney, NSW, Australia. His main research interests include machine learning, data analytics, Bayesian learning and inference, and image analysis.
\end{IEEEbiography}
\vskip -15pt plus -1fil
\begin{IEEEbiography}[{\includegraphics[width=1in,height=1.25in,clip,keepaspectratio]{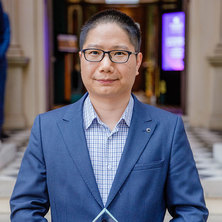}}]{Hongzhi Yin}
received his PhD degree in computer science from Peking University, in 2014. He is an associate professor and ARC Future Fellow with The University of Queensland. He received the Australian Research Council Discovery Early Career Researcher Award in 2016 and UQ Foundation Research Excellence Award in 2019. His research interests include recommendation system, user profiling, topic models, deep learning, social media mining, and location-based services.
\end{IEEEbiography}

\end{document}